\def\year{2022}\relax
\documentclass[letterpaper]{article} 
\usepackage{aaai22}  
\usepackage{times}  
\usepackage{helvet}  
\usepackage{courier}  
\usepackage[hyphens]{url}  
\usepackage{graphicx} 
\usepackage[dvipsnames]{xcolor}
\usepackage{natbib}  
\usepackage{caption} 
\DeclareCaptionStyle{ruled}{labelfont=normalfont,labelsep=colon,strut=off} 
\usepackage[ruled]{algorithm2e}
\usepackage{algorithmic}
\usepackage{multirow}

\usepackage{times}
\usepackage{epsfig}
\usepackage{graphicx}
\usepackage{amsmath}
\usepackage{amssymb}
\usepackage{enumitem}
\usepackage[font=small]{subcaption}
\usepackage{cleveref}

\title{Decompose the Sounds and Pixels, Recompose the Events}

\author{
    Varshanth R. Rao \textsuperscript{\rm 1},
    Md Ibrahim Khalil \textsuperscript{\rm 1,2},
    Haoda Li \textsuperscript{\rm 1,3},
    Peng Dai \textsuperscript{\rm 1 \footnote{Corresponding author.}},
    Juwei Lu \textsuperscript{\rm 1}
}
\affiliations{
    \textsuperscript{\rm 1}Huawei Noah's Ark Lab\\
    \textsuperscript{\rm 2}University of Waterloo, Canada\\
    \textsuperscript{\rm 3}University of Toronto, Canada\\

    \{varshanth.rao1, md.ibrahim.khalil, haoda.li, peng.dai, juwei.lu\}@huawei.com
}

\usepackage{bibentry}

\begin{document}

\maketitle

\begin{abstract}
In this paper, we propose a framework centering around a novel architecture called the Event Decomposition Recomposition Network (EDRNet) to tackle the Audio-Visual Event (AVE) localization problem in the supervised and weakly supervised settings. AVEs in the real world exhibit common unravelling patterns (termed as \emph{Event Progress Checkpoints} (EPC)), which humans can perceive through the cooperation of their auditory and visual senses. Unlike earlier methods which attempt to recognize entire event sequences, the EDRNet models EPCs and inter-EPC relationships using stacked temporal convolutions. Based on the postulation that EPC representations are theoretically consistent for an event category, we introduce the State Machine Based Video Fusion, a novel augmentation technique that blends source videos using different EPC template sequences. Additionally,  we design a new loss function called the Land-Shore-Sea loss to compactify continuous foreground and background representations. Lastly, to alleviate the issue of confusing events during weak supervision, we propose a prediction stabilization method called Bag to Instance Label Correction. Experiments on the AVE dataset show that our collective framework outperforms the state-of-the-art by a sizable margin.
\end{abstract}

\section{Introduction}

\label{section:intro}

As videos transform the landscape of information exchange and entertainment, audio-visual understanding becomes vital to augment the related applications. True to the quote by Harry Houdini - \textit{``What the eyes see and the ears hear, the mind believes"}, the successful perception of videos is powered through the activation and interplay of the visual and auditory sensory streams. A pragmatic testimony of the necessity of the audio-visual integration is the McGurk effect \cite{mcgurk1976hearing} in speech perception \cite{schwartz2002audio}. Additionally, ambient noise in real life like traffic or wind may mask or overlap with the sound of interest, rendering audio to be the noisier modality. The marriage of the two modalities has benefited various audio-visual tasks such as sound source localization and separation \cite{zhao2018sound, owens2018audio, NIPS2018_8002, arandjelovic2018objects, senocak2018learning}, audio-visual speech recognition \cite{noda2015audio, afouras2018deep}, etc. 

Under the umbrella of video understanding, the Audio-Visual Event Localization (AVEL) problem \cite{tian2018ave} has garnered increasing attention. AVEL entails the localization and identification of events which are both audible and visible. Supervised Event Localization (SEL) is to perform AVEL when the segment-level labels are available for supervision while for the Weakly Supervised Event Localization (WSEL) task, only the video level labels are available. Although the accompanying AVE dataset \cite{tian2018ave} is size-wise small (4143 videos covering 28 event categories), it is currently the only dataset to offer segment level audio-visual labels, hence facilitating the SEL task.

\begin{figure}[t]
\begin{center}
   \includegraphics[width=1\linewidth]{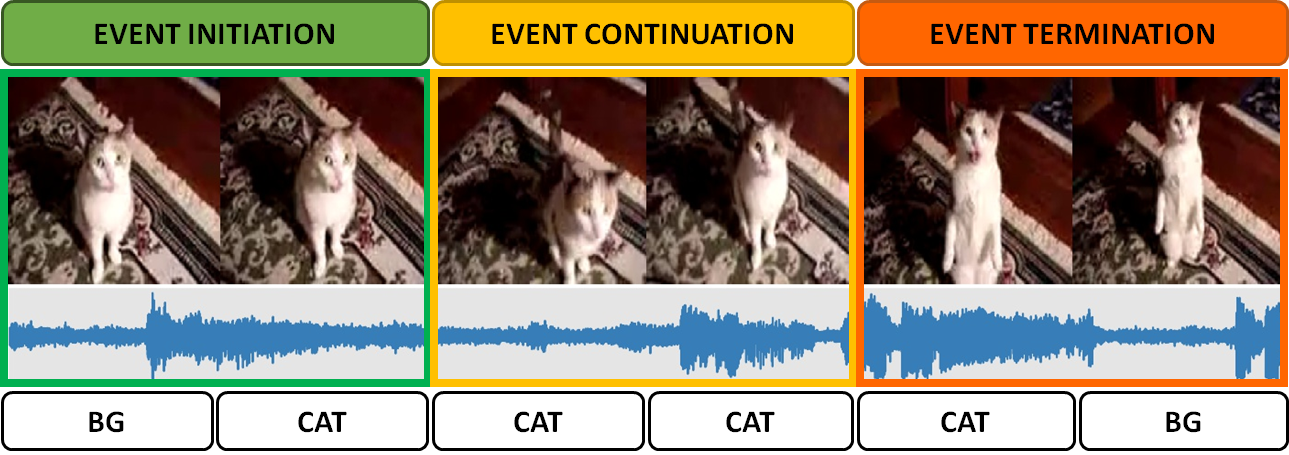}
\end{center}
\setlength{\belowcaptionskip}{-11pt}
   \caption{A depiction of event atomization. Events can be decomposed into Event Progress Checkpoints (top row) according to the temporal position along the event line. The corresponding audio-visual sequence is displayed in the middle row, while the ground truth segment labels are indicated in the bottom row.}
\label{fig:sub_event}
\end{figure}

Previous works resolve short and long-term temporal relationships by sequentially traversing through the video using memory networks like LSTMs. Digressing from their approach and inspired from the viewpoint of bottom-up temporal action localization \cite{lin2018bsn, lin2019bmn, liu2019multi}, we propose that events can be atomized into three Event Progress Checkpoints (EPCs); Event Initiation, Event Continuation, and Event Termination. We visualize these EPC formations in Fig. \ref{fig:sub_event}, where member segments delineate the commencement, uninterrupted procession, and completion of the event respectively. A natural choice for aggregating segments into EPCs is through Temporal Convolutional Networks (TCN) \cite{edtcn}. By stacking TCN layers, we can establish temporal dependencies (potentially longer than recurrent networks \cite{tcn}) at different granularities. Thus, we coalesce the fundamental concepts of the TCN and U-Net \cite{Ronneberger2015} to create a novel architecture called the Event Decomposition Recomposition Network (EDRNet). Leveraging EPCs (in place of segments) as the foundation units can lead to better generalization since an EPC representation of a category should be applicable across all videos of that category. Based on this assumption, we propose a State Machine Based Video Fusion technique which expands the current dataset by blending conceptually similar videos using random state-machine generated EPC sequences. 

Typically, AVEL is treated as a segment classification problem and supervision involves treating segments as independent units. As a consequence, the relationships within and between continuous foreground (FG) and background (BG) segments (called patches) are overlooked. Instead, we view patches as neighborhoods with internal (FG-FG/ BG-BG) similarities and external (FG-BG) differences. We strengthen these relationships by supervising the EDRNet with a new Land-Shore-Sea loss function. 

Prior works fail to explain the underlying cause of the performance gap between SEL and WSEL methods. In our investigation, we observe that WSEL methods suffer from a higher confusion between similar AVEs (such as helicopter vs airplane) due to smaller error gradients arising from weak supervision. To alleviate this issue, we extend a Multi-Instance Learning (MIL) method termed as the Bag to Instance Label Correction to refine FG localization predictions inside confident neighborhoods. In summary, our contributions can be highlighted as follows:

\begin{enumerate}
    \item Based on event atomization, we propose the Event Decomposition Recomposition Network (EDRNet), a novel TCN-based architecture to tackle the AVEL problem.
    \item We describe how the State Machine Based Video Fusion blends conceptually similar but content-wise different videos from segment annotated source videos, thus expanding the AVE dataset and fueling data-hungry deep networks.
    \item We describe a heuristic method, the Bag to Instance Label Correction, to rectify incorrect FG predictions amongst confident FG neighborhoods.
    \item Experimental results show that the EDRNet framework outperforms the state-of-the-art methods for both SEL and WSEL tasks on the AVE dataset.
\end{enumerate}


\section{Related Works}
\textbf{Audio-Visual Event Localization (AVEL):}
Event localization entails the identification of regions of an input sequence corresponding to the annotated event(s). Event localization has been classically viewed as an MIL problem \cite{mil}, where only the coarse level labels are available. Such works have explored the use of the visual modality in videos \cite{ved1, ved3}, audio modality in sound files \cite{sed2016, seld19}, and combinations of both along with video attributes and text in videos \cite{med1, med2}. Recently released, the AVE dataset \cite{tian2018ave} contains videos annotated both at the video and segment level for audible and visible events. \cite{lin_accv_2020} devise an Audiovisual Transformer to use audio as the guiding modality to refine visual features by performing spatial attention on contextual frames and instance attention to locate the sound-source within a frame. The Positive Sample Propagation module by \cite{Zhou_CVPR_2021} calculates similarity matrices between audio and visual features of different segments and thresholds them to eliminate insignificant audio-visual pairs. These matrices are used to co-refine similar segments together before fusing the modality information and learning temporal dependencies using LSTMs. Different from prior methods, we capture temporal dependencies progressively by stacking temporal convolutions. Further, in our qualitative analysis, we demonstrate that cross-modal guidance can be achieved \emph{implicitly} through modality fusion.

\textbf{Multi-modal Fusion:}
Multi-modal learning can result in more robust inferences since each modality can supply unique information relevant to the downstream task. Adopting the correct fusion technique becomes vital for exploiting each modality. Although prior works across different domains \cite{snoek2005early, affect_recogntion_early_vs_late, kim2019improving, alberti2019fusion} have explored and compared early and late stage fusion methods, it remains inconclusive which of the two is superior. In \cite{med1}, double fusion is proposed wherein late fusion is achieved by aggregating results of classifiers which operate on combinations of early fused features. Hybrid fusion is described in \cite{atrey2010multimodal} wherein the outputs of early and late fusion units are further merged to yield the final decision unit. Past AVEL methods \cite{tian2018ave, Ramaswamy_2020_ICASSP, Zhou_CVPR_2021} adopt LSTMs to fuse the audio and visual modalities while learning temporal alignments. In our work, the EDRNet executes \emph{dual-phase modality fusion} wherein an early fusion branch amasses crucial cross-modal dependencies and later incrementally refines individual modality representations.

\begin{figure*}
\begin{center}
\includegraphics[width=0.825\textwidth]{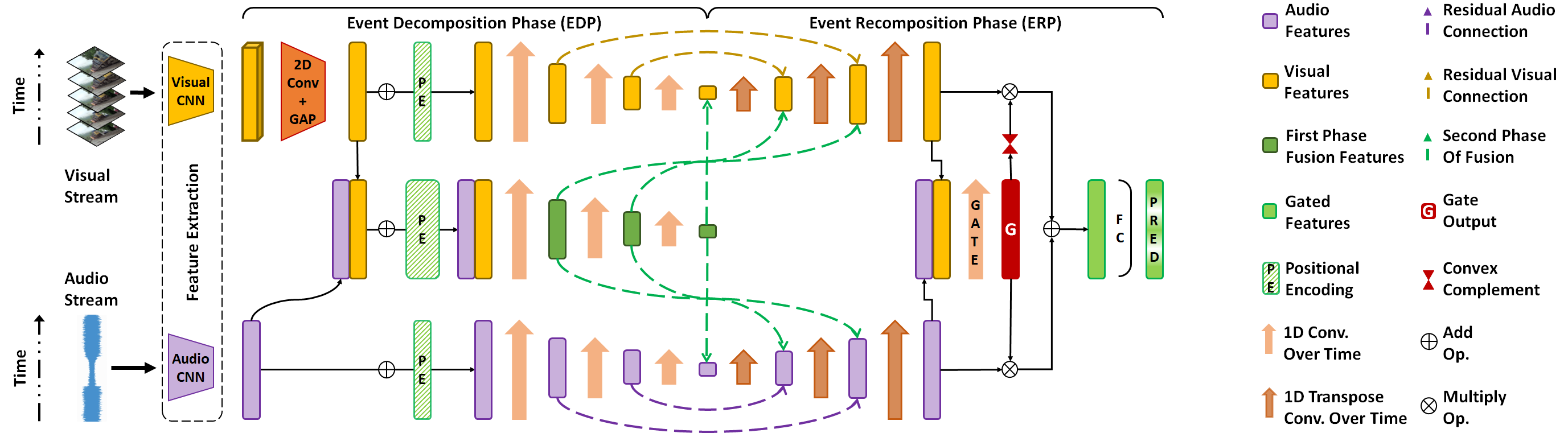}
\end{center}
\setlength{\belowcaptionskip}{-11pt}
   \caption{Overview of the Event Decomposition Recomposition Network. Modality-wise localization features are forged and refined in two phases: the Event Decomposition Phase (EDP) and the Event Recomposition Phase (ERP). The EDP summarizes the video into an event composition which the ERP leverages to effectively localize events in increasing temporal granularity. Consensus of the modalities is learned by a gating mechanism to yield the final audio-visual event localization predictions. 
   }
\label{fig:edrnet}
\end{figure*}

\section{Methodology}
\subsection{Problem Statement and Notations}
In the AVEL problem, each video sequence is split into $N$ non-overlapping segments. The segment level event label is denoted by $y_{t}=\{y_{t}^{c} | y_{t}^{c} \in \{0, 1\}, \sum_{c=0}^{C-1}y_{t}^{c}=1\}$ while the video level event label is denoted by $y=\{y^{c} | y^{c} \in \{0, 1\}, \sum_{c=0}^{C-1}y^{c}=1\}$. Here $C$ denotes the number of event classes inclusive of a BG event indicating independently audible (or visible) events or the absence of an event. For each video segment, the audio and visual features are extracted and denoted as $\{f_{t}^{A}, f_{t}^{V}\}_{t=1}^{N}$ respectively. Here $f_{t}^{A} \in \mathcal{R}^{d_a}$ and $f_{t}^{V} \in \mathcal{R}^{{d_v} \times S}$ where $d_a$ is the dimension of the audio features, $d_v$ and $S$ are the dimension and the spatial size of the visual feature maps respectively. Following \cite{tian2018ave}, we fix the feature extractors and build our architecture on top of these local features. SEL and WSEL tasks entail the prediction of the segment level event label $\hat{y_t}$, wherein $y_t$ is available to use for training in SEL and only the video level label $y$ is available for WSEL. 

\subsection{Event Decomposition Recomposition Network}
\label{subsection:edrnet}
The EDRNet, as depicted in Fig. \ref{fig:edrnet}, tackles both the SEL and WSEL tasks and operates in two phases; the Event Decomposition Phase (EDP) and the Event Recomposition Phase (ERP). The EDRNet utilizes the EDP to form a global understanding by encoding the video's event composition into temporally compressed audio, visual, and audio-visual representations. The audio and visual representations are then temporally upsampled in the ERP to ``recompose" the event localizations from each modality's perspective at the segment level. Residual connections from the EDP supplies audio, visual, and audio-visual cues to the ERP's modality-wise localization branches. We detail both phases below.

\smallskip
\noindent
\textbf{Event Decomposition Phase:}
Given $f_{t}^{A}$ and $f_{t}^{V}$, the EDP amasses information from fixed-sized segment sequences (decomposed events) into a global representation capable of describing the video's event composition. We extract the spatial information from $f_{t}^{V}$, using a 2D convolutional layer followed by a Global Average Pooling (GAP), yielding a condensed feature vector $\hat{f}_{t}^{V} \in \mathcal{R}^{d_{v}}$. To form composition perspectives, we build three independent modal branches; audio only, visual only and fused audio-visual, initialized as $D_{A}=\{f_{t}^{A}\}_{t=1}^{N}$, $D_{V}=\{\hat{f}_{t}^{V}\}_{t=1}^{N}$ and $D_{AV}=\{f_{t}^{AV}\}_{t=1}^{N}$ respectively where $f_{t}^{AV} = [f_{t}^{A}, \hat{f}_{t}^{V}]$ and $[.]$ is the concatenation operation. Following \cite{transformer}, we add positional encoding $p_t$ to each branch's initial features $f_{t}^{b} \in \mathcal{R}^{d_b}$, $b \in \{A, V, AV\}$ to inform the subsequent network components about the positional (temporal) relationship between the local features within each branch. Hence we define $p_{t}^{i}$ for all $i \in [1, d_{b}]$ and integer $k$ as:
\begin{equation}
p_{t}^{i} = 
\begin{cases}
    sin(\omega_{k} t), & \text{if } i=2k\\
    cos(\omega_{k} t), & \text{if } i=2k+1
\end{cases}
\text{where }
\omega_{k} = \frac{1}{10^{8k/{d_b}}}
\label{eq:positional_encoding}
\end{equation}

To learn event patterns at different temporal granularity, layers of temporal convolutions, which we term as decomposition operations, are employed in each modal branch. Specifically for each modal branch $b$, we define at layer $l_{dec}$, the output of a temporal convolution $\mathcal{F}$ with parameters $W_{b}^{l_{dec}}$  using kernel size $k$ as:
\begin{equation}
    D_{b}^{l_{dec}} = \mathcal{F}(D_{b}^{l_{dec}-1}, k; W_{b}^{l_{dec}})
    \text{ where }
    D_{b}^{0} = D_{b}
\label{eq:temporal_decomposition}
\end{equation}

After $L$ decomposition layers, we denote the modal branch outputs as $D_{b}^{L} \in \mathcal{R}^{N_{dec} \times {d_{L}}}$, where, $N_{dec}$ and $d_{L}$ are the temporal and feature map dimensions respectively.

\smallskip
\noindent
\textbf{Event Recomposition Phase:}
In the EDP, the audio-visual branch performs fusion of the constituent modalities to gather a joint perspective by forming cross-modal associations. EDP feature maps at layer $l_{dec}$, i.e., $D_{b}^{l_{dec}}$, contain both event specific positional information as well as event compositions local to the receptive field at layer $l_{dec}$. In the ERP, we use $D_{A}^{L}$ and $D_{V}^{L}$ as a foundation to forge progressive modality-wise localization branches using temporal transpose convolutions, which we term as recomposition operations.
Additionally, we derive modality-specific guidance and dissemination of cross-modal knowledge through residual connections from the corresponding EDP's modality-specific layers and fusion layers at the same receptive field $RF$. The cross-modal knowledge gained by early fusion in the EDP and its flow into the ERP is termed as \emph{dual-phase modality fusion}. It is noteworthy to highlight that without the inclusion of the positional encoding in the EDP, the decomposition operations gradually results in the loss of the events' position information within a video, which can be crucial for the ERP to form effective localization features. We formulate the recomposition output of branch $b' \in \{A, V\}$ at layer $l_{rec}$ using temporal transpose convolutions of kernel size $k$ and parameters $W_{b'}^{l_{rec}}$ as:
\begin{equation}
\begin{split}
    R_{b'}^{l_{rec}} = \mathcal{F}^{T}(R_{in}, {k}; W_{b'}^{l_{rec}})
&\\
    R_{in} =
    \begin{cases}
        D_{b'}^{L} + D_{AV}^{L}, & \text{if }{l_{rec}}=1,\\
        R_{b'}^{l_{rec}-1} + D_{b'}^{l_{dec}} + D_{AV}^{l_{dec}}, & \text{otherwise}
    \end{cases}
    &\\
\end{split}
\label{eq:temporal_recomposition}
\end{equation}

\noindent where $RF(l_{rec}) = RF(l_{dec})$. After $L$ layers of recomposition operations, we denote the recomposed modal branch outputs as $R_{b'}^{L} \in \mathcal{R}^{N \times d'_{L}}$. Here, $R_{A}^{L}$ and $R_{V}^{L}$ are features which respectively form the audio and visual perspectives of the event localizations. To successfully localize and identify AVEs, a consensus is required from both the modalities, the extent of which is context dependent. For example, to recognize a dog barking when its mouth is not clearly visible (weak visual signal), the model must seek out a barking sound in the audio modality. On the other hand, when its face is clearly visible but it is growling silently, the visual cue of the dog's slightly open mouth could aid in identifying the barking event. To capture this intricate cross-modal dependency, we learn a gating function through a temporal convolutional layer $\mathcal{F}_G$ with a sigmoid activation that operates on a fusion of both event localization perspectives. 
\begin{equation}
    G = \sigma({\mathcal{F}_{G}([R_{A}^{L}, R_{V}^{L}], k ; W_{G})}) \in \mathcal{R}^{N \times d'_{L}}
\end{equation}
where $[.]$ is the concatenation operation. We express the audio-visual perspective $R_{AV}^{G}$ as a convex combination of the audio and visual localization perspectives with the coefficients as the gated output $G$ and its complement respectively. We transform $R_{AV}^{G}$ into localization predictions over $C$ categories using an FC followed by a softmax activation. 
\begin{equation}
    R_{AV}^{G} = G \odot R_{A}^{L} + (1-G)\odot R_{V}^{L}
\end{equation}
\begin{equation}
    \hat{y_t} = \mathrm{Softmax}(W_{seg} R_{AV}^{G} + b_{seg})
\label{eq:localization_pred}
\end{equation}

\noindent
where denotes $\odot$ the Hadamard Product. For the SEL task, we supervise the segment level predictions using the multi-class cross entropy loss $\mathcal{L}_{seg} = CE(y_t, \hat{y_t})$. The WSEL is formulated as an MIL problem where the video and segment level labels represent the bag and instance labels respectively. We use MIL pooling to aggregate the segment predictions into a single video level prediction $\hat{y}$.
\begin{equation}\label{eq:wsel_pred}
    \hat{y} = \frac{1}{N} \sum_{t=1}^{N} \hat{y_t}, \text{   } \hat{y} \in \mathcal{R}^C
\end{equation}

\subsection{Land-Shore-Sea Loss}

While $\mathcal{L}_{seg}$ can guide the EDRNet to classify individual segments correctly, it cannot capture the positive correlation within continuous FG (or BG) segments (called patches) and the negative correlation between the FG and BG patches. Ideally, FG features of the same event type should be closer while being further away from BG features. However, FG segments may not be discretely distinguishable from BG ones due to the following obstacles: 
\begin{enumerate}[label=O\arabic*, itemsep=0.25ex, topsep=3pt]
    \item \label{lss_motiv_1} A segment's audio/visual/audio-visual cues may be weak (or strong) leading to its features to be undesirably closer to that of the BG (or FG) event.
    \item \label{lss_motiv_2} FG events at the border may not span the entirety of the segment length and hence may be annotated as BG.
\end{enumerate}

\begin{figure}[t]
\begin{center}
   \includegraphics[width=1\linewidth]{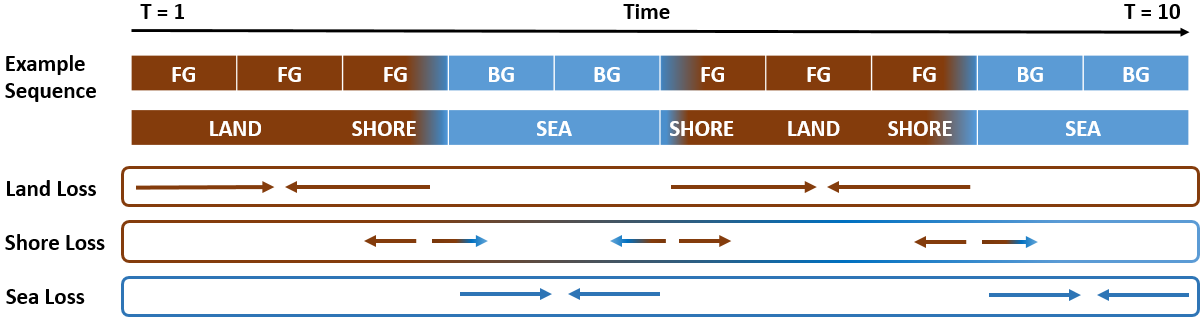}
\end{center}
\setlength{\belowcaptionskip}{-11pt}
   \caption{Visualization of the land, shore and sea losses. The land and sea losses aim to bring closer the features of continuous FG and BG segments respectively. The shore loss aims to pull closer the features of a FG border to that of neighboring FG segments (Land) while pushing it away from that of the neighboring BG segments (Sea).}
\label{fig:lssloss}
\end{figure}

Motivated from natural topography, we view the FG patches as ``Lands", BG patches as ``Seas" and the FG event borders as the ``Shores". Intuitively, a patch's representation should be a generalization of its constituent segments. If a majority of the segments in the patch exhibit strong audio-visual cues toward the correct event, the patch representation could be a beacon for its constituents. The Land (or Sea) loss hones the features of a FG (or BG) segment exhibiting \ref{lss_motiv_1} by bringing it closer to that of the Land (or Sea). Similarly, at the shore where segments can exhibit \ref{lss_motiv_1} and \ref{lss_motiv_2}, the Shore loss draws the features of the event border (Shore) nearer to that of the FG patch (Land) while driving them apart from that of the BG patch (Sea). We depict the mechanism of each loss in Fig. \ref{fig:lssloss}. Let the gated features $R_{AV}^{G}$ be denoted as $L$, $S$ and $Sh$ for land, sea and shore respectively. For a video sample $i$, we denote $N_{l}^{i}$, $N_{s}^{i}$ and $N_{sh}^{i}$ as the number of lands, seas and shores respectively. The land and sea loss are achieved by minimizing the $L_{2}$ distance between the average feature representations of the first half and second half of the land and sea patch respectively:
\begin{equation}
    \mathcal{L}_{land}^{i} = \frac{1}{N_{l}^{i}}\sum_{l=1}^{N_{l}^{i}} \lvert \lvert \overline{L_{l}^{1st}} - \overline{L_{l}^{2nd}} \rvert \rvert_{2}
\end{equation}
\begin{equation}
    \mathcal{L}_{sea}^{i} = \frac{1}{N_{s}^{i}}\sum_{s=1}^{N_{s}^{i}} \lvert \lvert \overline{S_{s}^{1st}} - \overline{S_{s}^{2nd}} \rvert \rvert_{2}
\end{equation}

The triplet loss is poised to implement the shore loss. The features of the shore represent the anchor sample while the average of the features of neighboring land and sea represent the positive and negative samples respectively:
\begin{equation}
    \mathcal{L}_{shore}^{i} = \frac{1}{N_{sh}^{i}}\sum_{s'=1}^{N_{sh}^{i}} \Big[\lvert \lvert Sh_{s'} - \overline{L_{s'}} \rvert \rvert_{2} - 
    \lvert \lvert Sh_{s'} - \overline{S_{s'}} \rvert \rvert_{2} + \alpha \Big]_{+}
\end{equation}
where $[x]_{+} = max(0,x)$ and $\alpha$ is the minimum margin to be maintained between $\lvert \lvert Sh_{s'} - \overline{L_{s'}} \rvert \rvert_{2}$ and $\lvert \lvert Sh_{s'} - \overline{S_{s'}} \rvert \rvert_{2}$. Thus, we define the LSS loss as $\mathcal{L}_{lss} = \mathcal{L}_{land} + \mathcal{L}_{sea} + \mathcal{L}_{shore}$ and train the EDRNet for the SEL task with the overall objective function as $\mathcal{L}_{SEL} = \lambda_{1} \mathcal{L}_{seg} + \lambda_{2}\mathcal{L}_{lss}$

\subsection{State Machine Based Video Fusion}

We exhibited that events can be atomized into 3 Event Progress Checkpoints (EPCs): Event Initiation (EI), Continuation (EC), and Termination (ET). EPCs can be further dissected based on the minimum temporal length required for the EPC content to unfold. We denote BG sequences, EI, EC and ETs of length $L \in \{1,2\}$ as BG\_$L$, START\_$L$, CONTINUE\_$L$ and END\_$L$ respectively. Table \ref{states} contains the search and extract segment patterns for all $\langle \text{EPC/BG} \rangle\_L$.

\begin{figure}[t]
\begin{center}
   \includegraphics[width=.48\textwidth]{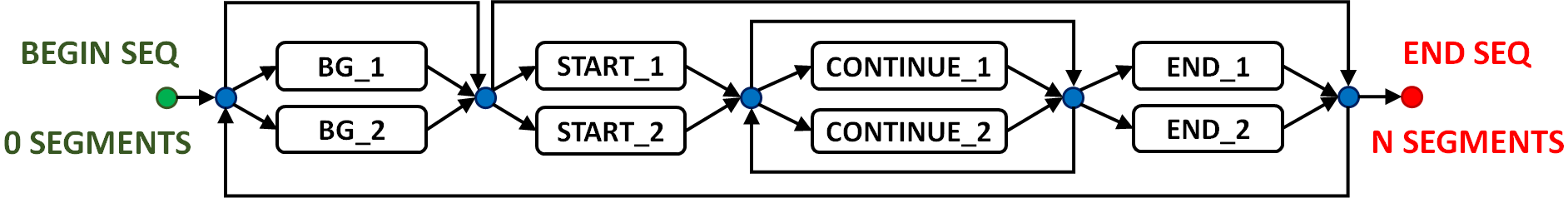}
\end{center}
\setlength{\belowcaptionskip}{-11pt}
\caption{State machine used in the SMB Video Fusion technique. The state machine generates an N-segment state-sequence by following the displayed rules of state transition.}
\label{fig:statemachine}
\end{figure}

The samples extracted for an EPC from videos of a particular category should reflect similar semantic content. For example, the EI sequence for any baby crying usually involves the baby's eyes crunching up and the opening of the mouth, followed by the shrill sound of the cry. Thus, by replacing EPC-specific audio-visual content with other samples of the same EPC, we can synthesize an entirely new video expressing the same semantic event sequence. This allows localization models to focus on identifying the core conceptual audio-visual correlations describing an EPC. However, this approach does not alter the sequence of EPCs and therefore limits the augmentation to the same sequence fingerprint. To introduce variation at the sequence level, we need to create different EPC blends. By considering all $\langle \text{EPC/BG}\rangle\_L$s as states, we model the blend synthesis process using a state machine (SM) (shown in Fig. \ref{fig:statemachine}) since it must adhere to the positional constraints of EPCs.

\begin{table}
\centering
\scalebox{0.85}{
\begin{tabular}{ |c|c|c| }
 \hline
 State & Segment Pattern & State Length\\
 \hline
 BG\_1 & [BG] & 1 Segment\\
 BG\_2 & [BG, BG] & 2 Segments\\
 START\_1 & [FG], Last N-1 Segments & 1 Segment\\
 START\_2 & [BG, FG] & 2 Segments\\
 CONTINUE\_1 & FG, [FG], FG & 1 Segment\\
 CONTINUE\_2 & FG, [FG, FG], FG & 2 Segments\\
 END\_1 & First N-1 Segments, [FG] & 1 Segment\\
 END\_2 & [FG, BG] & 2 Segments\\
 \hline
\end{tabular}}
\caption{Search \& extraction segment patterns for all states. Within N-segment videos, we search for the complete segment pattern and extract only the ones enclosed within square brackets to form an instance for the state.}
\label{states}
\end{table}

\begin{figure}[h]
\begin{center}
\includegraphics[width=0.45\textwidth]{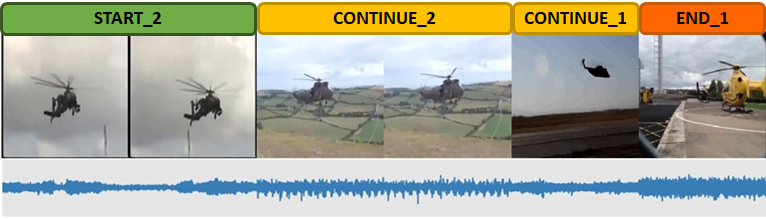}
\end{center}
\setlength{\belowcaptionskip}{-11pt}
   \caption{SMB fused video snippet of a ``Helicopter" event. On top is the state sequence generated by the state machine. The bottom are the corresponding state-based clips which are stitched together from different helicopter videos.
   }
\label{fig:smb_example}
\end{figure}

Using Algorithm \ref{smb_video_fusion_algo}, we combine the above two strategies into a novel video augmentation technique called as the State Machine Based Video Fusion. Fig. \ref{fig:smb_example} illustrates a snippet of an example output with the corresponding state sequence for the ``Helicopter" event. As seen, the produced sequence is a coherently structured amalgamation of the source videos. 

\begin{algorithm}
\SetAlgoLined
\SetNlSty{texttt}{}{:}
\SetKwInOut{Input}{Output}
\KwIn{Training set videos of event type $e$ and length $N$, Number of output videos $N_v$}
\KwOut{$N_v$ fused videos of event type $e$}
 \nl Initialize a database for each state\\
 \nl Identify the states for each video using Table \ref{states}\\
 \nl Store state specific video content into respective state databases\\
 \nl \For {$i \gets 1$ to $N_v$}  {
  \nl Using the SM, generate a $N$-segment state sequence $SEQ_{state} = \{s_1, s_2, ..., s_{N_{state}}$\}\\
  \nl Initialize output video $V_i$\\
  \nl \For{$j \gets 1$ to $N_{state}$}  {
  \nl Choose a random sample $v$ from $s_j$-database\\
  \nl Append $v$ to $V_i$
 }
 }
\caption{SMB Intra-Class Video Fusion}
\label{smb_video_fusion_algo}
\end{algorithm}

\subsection{Bag to Instance Label Correction}
\begin{figure}[t]
\begin{center}
   \includegraphics[width=1\linewidth]{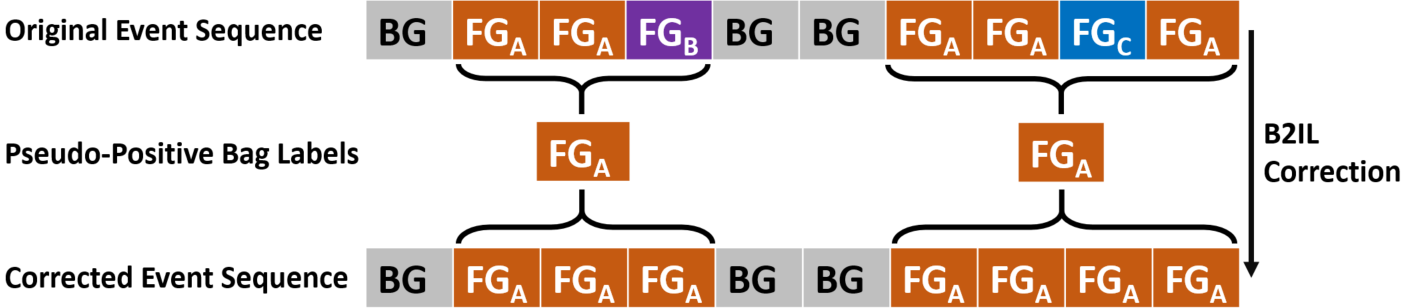}
\end{center}
\setlength{\belowcaptionskip}{-11pt}
   \caption{An example of B2ILC where incorrect $\text{FG}_\text{B}$ and $\text{FG}_\text{C}$ instances found in two pseudo-positive bags are corrected as $\text{FG}_\text{A}$.}
\label{fig:b2il}
\end{figure}

From an MIL perspective of a localization sequence, a positive bag is usually of an FG event type, implying that negative instances can source from other FG types as well the BG. We tackle both sources separately by encapsulating continuous correct and incorrect FG predictions into a single pseudo-positive set. Discerning between BG and FG instances requires comprehension of video content which we entrust the model to perform. We can curtail the intra-FG event confusion within pseudo-positive sets by using the following MIL approach. In MIL, Witness Rate (WR) is the proportion of positive instances in the bag. When the WR is sufficiently high, the instance labels can be assumed to be that of the bag \cite{Carbonneau_2018}. We term this process as ``Bag to Instance Label Correction (B2ILC)".
During inference time, we assume that a sufficiently trained localization model possesses a high FG (pseudo-positive label) precision, and hence we can treat continuous FG segment predictions to be a pseudo-positive bag. By setting the WR threshold as 0.5, the B2ILC becomes a strict majority voting process and on application to the pseudo-positive bag, all constituent instances (segments) would inherit the dominant FG label. We note that different from pure majority voting, B2ILC is constrained by the WR threshold. Fig. \ref{fig:b2il} demonstrates B2ILC with an example event sequence.

\section{Experiments and Results}
\smallskip
\noindent
\textbf{Dataset and Evaluation Metrics:} The AVE dataset \cite{tian2018ave} is a subset of the AudioSet \cite{audioset} containing 4143 videos, each 10 seconds long, i.e., $N=10$. There are 28 (+1 for BG) diverse event classes covering vehicle sounds, animal activity, instrument performances, etc. Video and segment level labels are available with clearly demarcated temporal boundaries. Events must be both audible and visible and spans for at least two seconds. We adopt the same train/validation/test split as \cite{tian2018ave}. Recently, \cite{Zhou_CVPR_2021} corrected the annotations for some test videos and report their performance on their method on this corrected test set. We denote the AVE dataset with the \emph{original} test set as \textbf{O-AVE}, and that with the \emph{corrected} test set as \textbf{C-AVE}. Following prior works, we evaluate the localization performance using the global classification accuracy of segment level predictions.

\smallskip
\noindent
\textbf{Implementation Details:} For a fair comparison with prior works, we utilize the same extracted audio and visual features (provided with the AVE dataset) using VGGish \cite{vggish} and VGG19 \cite{vgg} networks pretrained on AudioSet \cite{audioset} and ImageNet \cite{imagenet} respectively. We configure the EDRNet with $k=3$, $L=4$, and a network width $d_l = 768$ for all layers. Sourcing the training set, we generate 250 samples per category using SMB video fusion. The optimization parameters for training EDRNet are specified in the \emph{Supplementary Material}. Hyperparameter tuning was performed using the validation set. 

\subsection{Benchmarking against SoTAs}
In Table \ref{tbl:perf_sotas}, we compare our EDRNet framework to the Audiovisual Transformer (AVT) \cite{lin_accv_2020} and Positive Sample Propagation (PSP) Network \cite{Zhou_CVPR_2021} on the SEL and WSEL tasks on O-AVE and C-AVE datasets. Our EDRNet framework which leverages TCNs to progressively recognize EPCs, outperforms the PSP on the O-AVE, by \textbf{1.06\%} and \textbf{1.09\%} on the WSEL and SEL tasks respectively. On the C-AVE, it outperforms PSP by \textbf{1.07\%} and \textbf{1.11\%} on the WSEL and SEL tasks respectively.

\begin{table}[]
\centering
\scalebox{0.85}{
\begin{tabular}{c|c|c|c}
\hline
\multirow{2}{*}{AVEL Method} & \multirow{2}{*}{Dataset} & WSEL & SEL \\
& & Acc (\%) & Acc (\%) \\
\hline
AVT \cite{lin_accv_2020} & O-AVE & 70.20 & 76.80  \\
PSP \cite{Zhou_CVPR_2021} & O-AVE & 72.80 & 76.84 \\
PSP \cite{Zhou_CVPR_2021} & C-AVE & 73.50 & 77.80 \\
\hline
\textbf{EDRNet (Ours)} & \textbf{O-AVE} & \textbf{73.86} & \textbf{77.93} \\
\textbf{EDRNet (Ours)} & \textbf{C-AVE} & \textbf{74.57} & \textbf{78.91} \\
\hline
\end{tabular}}
\setlength{\belowcaptionskip}{-11pt}
\caption{Performance comparison with SoTAs (\%) for the SEL and WSEL task on the O-AVE and C-AVE datasets.}
\label{tbl:perf_sotas}
\end{table}

\subsection{Ablation Studies}
We report our ablation studies only on O-AVE i.e. the AVE dataset with the \emph{original} test set.

\smallskip
\noindent \textbf{Framework Decoupling:}
We investigate the contribution of each component and summarize the results for the SEL and WSEL tasks in Table \ref{tbl:SEL_WSEL_ablation}. The EDRNet fueled by the SMB video fusion proves to be highly effective, contributing to a \textbf{0.83\%}  increase in SEL accuracy. It bolsters the performance of challenging categories such as ``Bus" (+19\%) and ``Female Speech" (+8\%) where conceptual focus is critical to wane off ambient distractions such as traffic and presence of other humans for the former and latter respectively.  The success of the EDRNet and the SMB video fusion corroborates the efficacy of EPC-based methods. The LSS loss contributes to a \textbf{0.52\%} increase for the SEL task and in the \emph{Supplementary Material}, we show that it compactifies the segment features within an event type. B2ILC benefits both tasks, but it influences the WSEL more. For WSEL, the MIL pooling (Eqn. \eqref{eq:wsel_pred}) ensures equal gradient distribution across all segments. Consequently, the difficult segments of commonly confused events (such as ``Truck" vs ``Bus", ``Motorcycle" vs ``Racecar",  etc.) cannot get prioritized. However, the resolution of easier segments increases the WR of pseudo-positive bags, thereby permitting B2ILC to disambiguate these hard positives (+10.5\% and +10\% for ``Truck" and ``Motorcycle" respectively).

\begin{table}[]
\centering
\scalebox{0.725}{
\begin{tabular}{c|c|c|c||c|c}
\hline
EDRNet & B2ILC & $\mathcal{L}_{lss}$ & SMBVF & SEL Acc. (\%) & WSEL Acc. (\%) \\ 
\hline
\hline
\checkmark & & & & 76.10 & 72.98 \\
\checkmark & \checkmark & & & 76.58 & 73.86 \\
\checkmark & \checkmark & \checkmark & & 77.10 & N/A \\
\checkmark & \checkmark & \checkmark  &\checkmark & 77.93 & N/A\\
\hline
\end{tabular}}
\setlength{\belowcaptionskip}{-11pt}
\caption{Component-wise ablation study of the EDRNet framework for the SEL and WSEL tasks. SMBVF indicates SMB Video Fusion and B2ILC indicates B2IL Correction.}
\label{tbl:SEL_WSEL_ablation}
\end{table}

\smallskip
\noindent
\textbf{EDRNet Configurations:} We study the influence of the 3 dimensions of EDRNet, i.e., temporal kernel size $k$, number of (de/re)composition layers $L$ and network width $d$, on the SEL task. For decompositions, we use a 1D temporal convolution of size $k$, unit stride and no padding. When applied on a sequence of length $N_{l}$ at layer $l \in [0, L_{max}]$, the output feature map will be of length $N'_{l+1} = N_{l} - k + 1$ where $N_{0}=N(=10)$. $L_{max}$ denotes the layer where the net receptive field is maximum, i.e., when $0 < N'_{L_{max}} < k$.

\begin{figure}[t!]
    \centering
    \begin{subfigure}[t]{0.245\textwidth}
        
        \includegraphics[height=0.95in,width=1.7in]{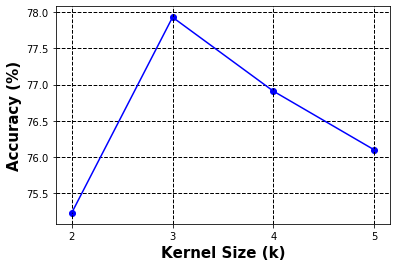}
        \caption{}
        \label{vary_k}
    \end{subfigure}%
     ~
    \begin{subfigure}[t]{0.245\textwidth}
        
        \includegraphics[height=0.95in,width=1.7in]{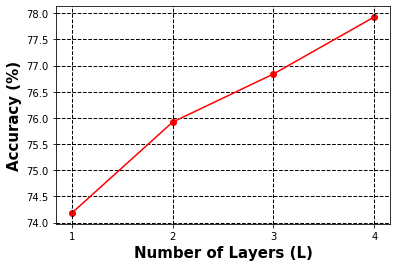}
        \caption{}
        \label{vary_L}
    \end{subfigure}
    
    \begin{subfigure}[t]{0.245\textwidth}
        
        \includegraphics[height=0.95in,width=1.7in]{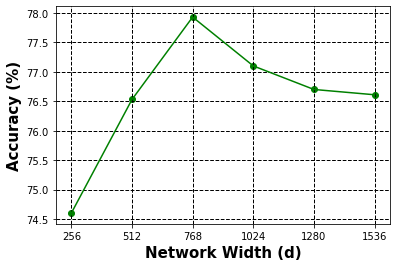}
        \caption{}
        \label{vary_d}
    \end{subfigure}%
     ~
    \begin{subfigure}[t]{0.245\textwidth}
        
        \includegraphics[height=0.95in,width=1.7in]{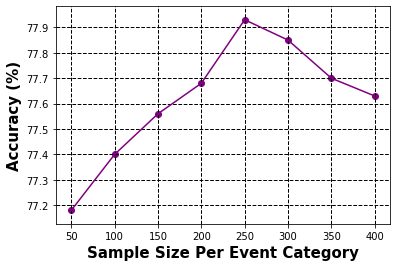}
        \caption{}
        \label{vary_sample_size_smb_fusion}
    \end{subfigure}

    \caption{For the SEL task, effect of (a)-(c): Varying the EDRNet's temporal kernel size $k$, number of (de/re)composition layers $L$ and network width $d$ resp. (d): Varying the extent of SMB video fusion.}
\end{figure}

We vary $k$ in the range $[2,5]$ and correspondingly configure the $L$ as $L_{max}^{k = \{2,3,4,5\}} = \{9, 4, 3, 2\}$ to induce maximum receptive fields (MRFs) for fair comparison. From Fig. \ref{vary_k}, it is evident that $k=3$ with $L_{max}^{k=3}=4$ is the optimal kernel size at the MRF. From the perspective of the first layer, the EDRNet makes a localization decision by considering the previous, current and next segment. Next, we fix $k=3$ and $d=768$ and vary $L$ in the range $[1,L_{max}^{k=3}]$ where $L_{max}^{k=3}=4$, to observe how increasing the net receptive field affects performance. From Fig. \ref{vary_L}, we discern that the network performs best at the MRF. Lastly, we freeze $k=3$ and $L=L_{max}^{k=3}=4$ and vary $d$ in the range $[256,1536]$ in steps of 256 to inspect the impact of network width. Fig. \ref{vary_d} shows that increasing $d$ brings benefit until the optimal value ($d=768)$, post which the network starts to overfit.

\smallskip
\noindent
\textbf{Impact of SMB Video Fusion:}
We investigate the effectiveness of the SMB video fusion by quantifying the gains upon varying the extent of augmentation. To produce balanced datasets, we generate $S$ samples per event type where $S$ is varied in the range [50, 400] in steps of 50. The SEL performances by the EDRNet trained on each of the augmented datasets are shown in Fig. \ref{vary_sample_size_smb_fusion}. The progressive inclusion of SMB fused videos aids the EDRNet till a certain limit (here, $S=250$), after which additional videos exposes the network to the same video clips, causing it to overfit.

\smallskip
\noindent
\textbf{Modal Branch Isolation Experiments:}
The clear demarcation of modal branches in the EDRNet permits the examination of branch level performances. Since the EDRNet gated output $R_{AV}^G$ is a convex combination of $R_{A}^{L}$ and $R_{V}^{L}$, they all share the same feature space. We can extract predictions from the audio (A) and visual (V) branches respectively by replacing $R_{AV}^{G}$ with $R_{A}^{L}$ and $R_{V}^{L}$ in Eqn. \eqref{eq:localization_pred}. Training various combinations of decomposition (DC) modal branches enables us to study the response of the activated recomposition (RC) branches. We visualize our study for better clarity in the \emph{Supplementary Material} and summarize our results on the SEL task in Table \ref{tbl:modality_isolation_result}. By enabling only the A (or V) DC branch (rows 1 \& 2), the corresponding RC output delivers the baseline performance for A (or V). The inclusion of the AV branch to DC A \& V separately (rows 3 \& 4) visibly boosts the RC A \& V performances due to the triggering of dual-phase modality fusion. Late fusion achieved by enabling both DC A \& V and disabling AV (row 5), allows each modal pathway to learn exclusive cues (reflected by low A \& V performances) to cover the counterpart's weaknesses and deliver a high gated accuracy. Finally, by training with all DC branches (row 6), the A \& V pathways learn sufficient exclusive cues to furnish the best gated accuracy. 

\begin{table}
\centering
\scalebox{0.725}{
\begin{tabular}{c|c|c|c||c|c|c}
 \hline
 \multicolumn{3}{c|}{Active DC Branch} &
 \multirow{2}{*}{Training Configuration} &
 \multicolumn{3}{c}{RC Branch Acc. (\%)} \\\cline{1-3}\cline{5-7}
 A & V & AV & & A & V & Gated \\
 \hline
 \hline
 \checkmark & - & - & A - Only & 65.84 & - & - \\
 - & \checkmark & - & V - Only & - & 66.93 & - \\
 \checkmark & - & \checkmark & A + Dual-Phase Fusion & 76.11 & - & - \\
- & \checkmark & \checkmark & V + Dual-Phase Fusion & - & 75.26 & - \\
\checkmark & \checkmark & - & A + V with Late Fusion & 54.38 & 62.76 & 75.88 \\
\checkmark & \checkmark & \checkmark & A + V + Dual-Phase Fusion  & 73.62 & 72.74 & 77.93 \\
\hline
\end{tabular}}
\setlength{\belowcaptionskip}{-11pt}
\caption{SEL performance of different modal branch configurations. We train different decomposition (DC) branches and measure performance from the corresponding activated recomposition (RC) branches. A, V and AV denote the audio, visual, and the DC audio-visual branch respectively.}
\label{tbl:modality_isolation_result}
\end{table}

\begin{figure}[h!]
\begin{center}
    \begin{subfigure}[t]{0.45\textwidth}
        \centering
        \includegraphics[width=1\textwidth]{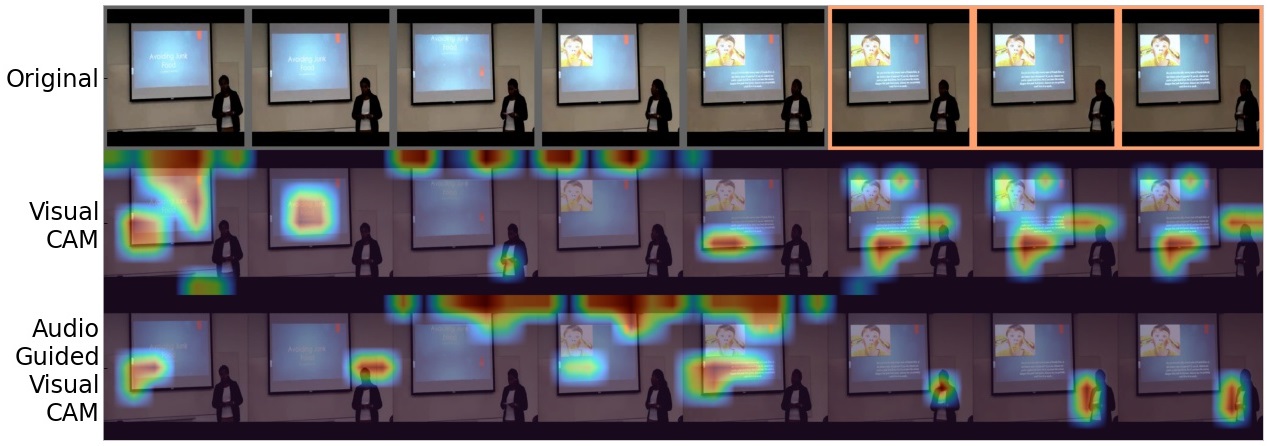}
        \caption{Audio guided visual focus illustrated on a ``Female Speech" video. Since the female is present in all segments, her visual presence is not discriminative enough to recognize the AVE. The focus shift to the female during her speech in the audio-guided visual CAM, indicates that the model leverages the audio to identify the AVE.}
        \label{CAM_agva}
    \end{subfigure}
    
 \begin{subfigure}[t]{0.45\textwidth}
        \centering
        \includegraphics[width=1\textwidth]{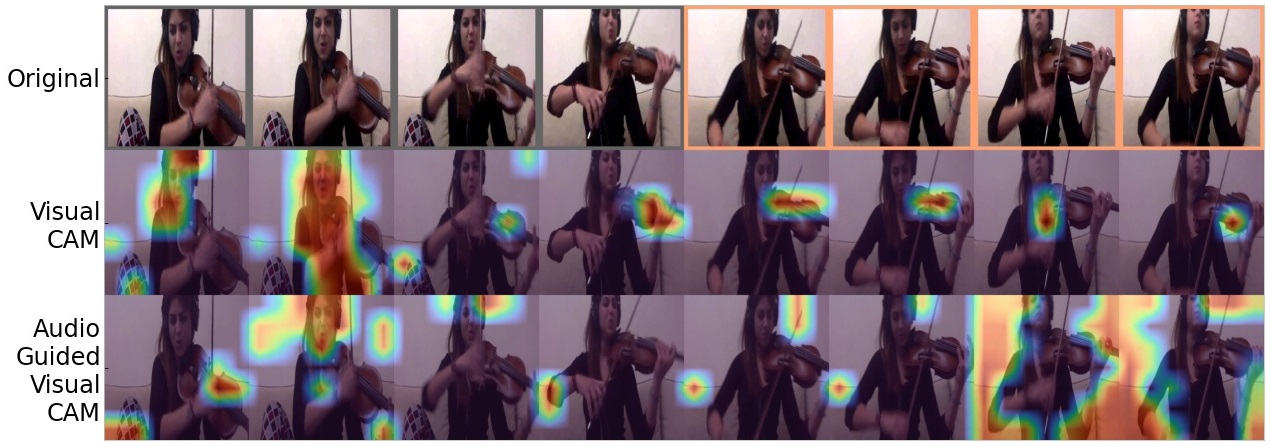}
        \caption{Independent visual focus demonstrated on a ``Violin Play" video. The visual CAM captures the model utilizing the visual cue of the contact of the violin bow on the violin to recognize the violin play AVE.}
        \label{CAM_vis_only}
    \end{subfigure}
\end{center}

\setlength{\belowcaptionskip}{-11pt}
   \caption{CAM visualizations from the EDRNet. Orange bordered segments indicate the presence of FG event.}
\label{fig:CAMs}
\end{figure}

\subsection{Qualitative Analysis}
\label{subsection:qualitative}

Our EDRNet framework outperforms prior works which leverage different attention mechanisms to perform and visualize AGVA. This raises an important question: do we need explicit attention mechanisms to achieve cross-modal guidance? To investigate, we use the kernel (sized $k$) weights corresponding to the maximum activation of $D_{V}^{1}$ and visual portion of $D_{AV}^{1}$ (refer Eqn \eqref{eq:temporal_decomposition}) as coefficients to the visual feature maps (FMs) corresponding to the GAP output $\hat{f}_{t}^V$. The $k$-averaged weighted sum of the FMs yields the Class Activation Maps (CAMs) for the visual and audio-guided visual branch respectively. For $k=3$, $D_{V}^{1}, D_{AV}^{1} \in \mathcal{R}^{8\times d_{1}}$ and hence 8 CAMs are generated for a 10 second video. Fig. \ref{fig:CAMs} presents CAMs for a ``Female Speech" and a ``Violin Play" video. In the former, when the woman speaks, the focus distinctly shifts to her in the audio guided visual CAMs. The network capitalized on the audio modality to attend on the visual modality, thereby \emph{implicitly} achieving AGVA. However, the sharp spatial focus on the violin from the visual CAMs in the latter suggests that the network is capable of exploiting only a single modality as well. The above alludes that the EDRNet accomplishes on-demand cross modal guidance without explicit attention mechanisms.
\section{Conclusion}

In this paper, we proposed the EDRNet to tackle the SEL and WSEL tasks. Unique to the EDRNet is the ability to localize audio, visual, and audio-visual events simultaneously through the structural assembly of individual and combined viewpoints. Propelled by event atomization, the SMB video fusion can augment datasets with semantically similar but spatiotemporally divergent videos. This exposes the model to diverse content and assists in calibrating its focus on the event source to tackle hard positives. The B2ILC heuristic leverages strong FG neighborhoods to stabilize the localization predictions. Our framework achieves state-of-the-art results on both tasks on the AVE dataset and the effectiveness of each module is validated through extensive experiments. 

\noindent
\textbf{Acknowledgements:} We thank Manasa Bharadwaj and Deepak Sridhar for the time and effort that they have put into carefully reviewing the various drafts of our paper.
\bibliography{aaai22}

\clearpage
\begin{figure}[h!]
\begin{center}
   \includegraphics[width=1\linewidth]{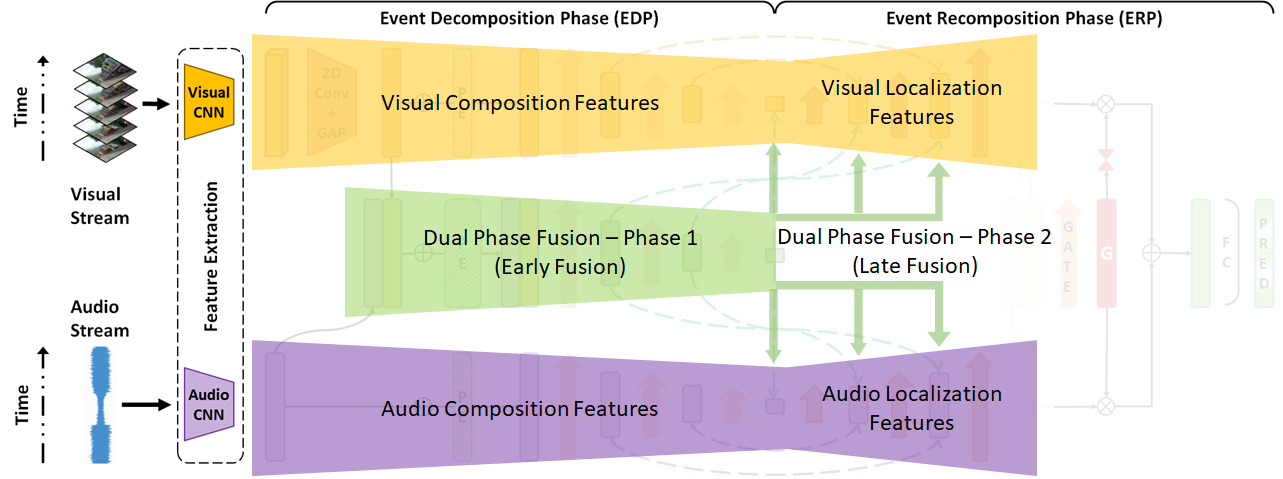}
\end{center}
   \caption{Visualization of the Dual-Phase Modality Fusion executed in the EDRNet. EDRNet architecture is elaborated in Section 4.1 and Figure \ref{fig:edrnet}}
\label{fig:edrnet_dual_phase_fusion}
\end{figure}

\section{Additional Experiments}
\subsection{Impact of the LSS Loss}
\label{subsection:lambda2}

\begin{figure*}[t!]
    \begin{subfigure}[t]{0.5\textwidth}
        \centering
        \includegraphics[width=1\textwidth]{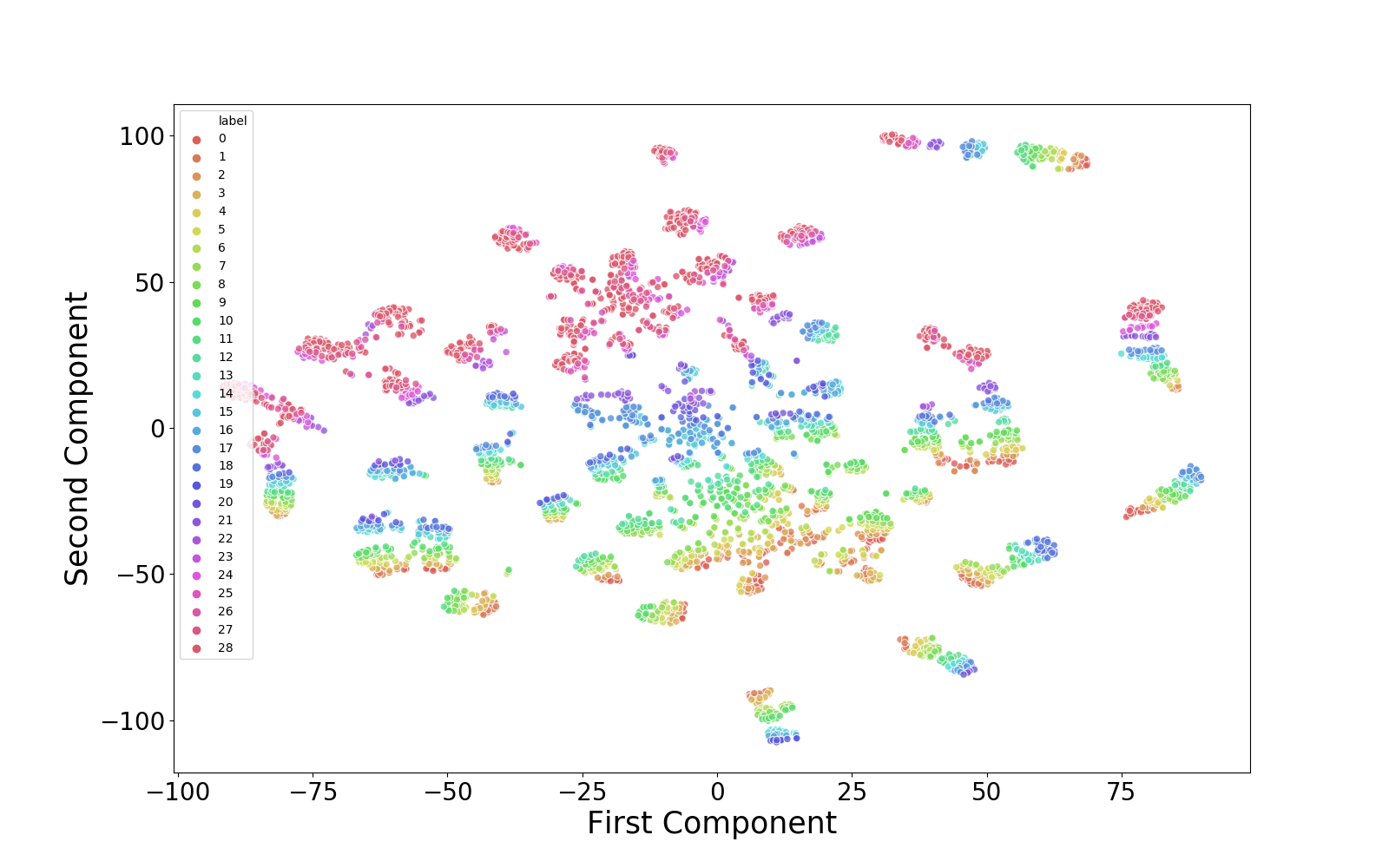}
        \caption{}
    \end{subfigure}%
     ~
    \begin{subfigure}[t]{0.5\textwidth}
        \centering
        \includegraphics[width=1\textwidth]{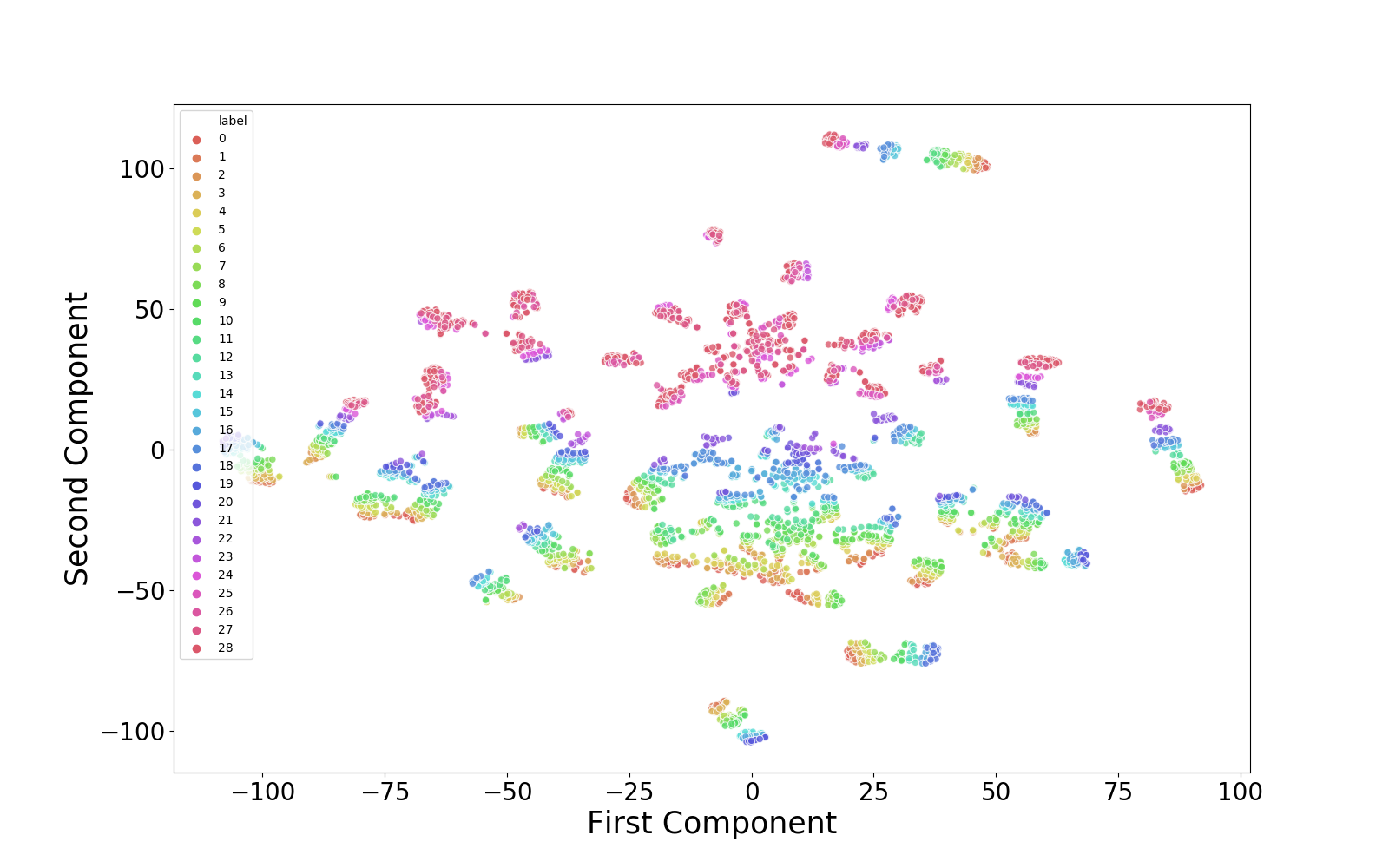}
        \caption{}
    \end{subfigure}
\caption{t-SNE plots for the EDRNet gated features $R_{AV}^{G}$ on the AVE dataset. (a) Without LSS Loss (b) With LSS Loss. Label 28 denotes the BG event. Apart from the constriction of feature representations, the demarcation between FG and BG representations become more distinct with the LSS loss.}
\label{fig:tsne_lss}
\end{figure*}

In the Methodology section, we explained the motivation and the mathematical formulation for the Land-Shore-Sea loss for supervising the EDRNet under the SEL setting. The Land and Sea loss aim to bring together the patch of continuous FG and BG segments respectively by reducing the distance of the representation of the 2 halves of the patch. Meanwhile, the Shore loss aims to bring closer the border FG segment (Shore) to the enclosing FG patch (Land) and push it away from the neighboring BG patch (Sea).
To capture the effect of the LSS loss on the segment features, we plot the EDRNet's gated features $R_{AV}^{G}$ with and without the LSS loss on the AVE dataset in Figure \ref{fig:tsne_lss}. We observe that with the LSS loss, the segment cluster representations move closer together to create compact event-wise groups. More importantly, the distinction between FG and BG (in red) clusters becomes more evident, hence alleviating the construction of an optimal separating hyperplane.

\subsection{Size Response of EDRNet Configurations}

\begin{figure*}[t!]
    \centering
    \begin{subfigure}[t]{0.3\textwidth}
        \centering
        \includegraphics[width=1\textwidth]{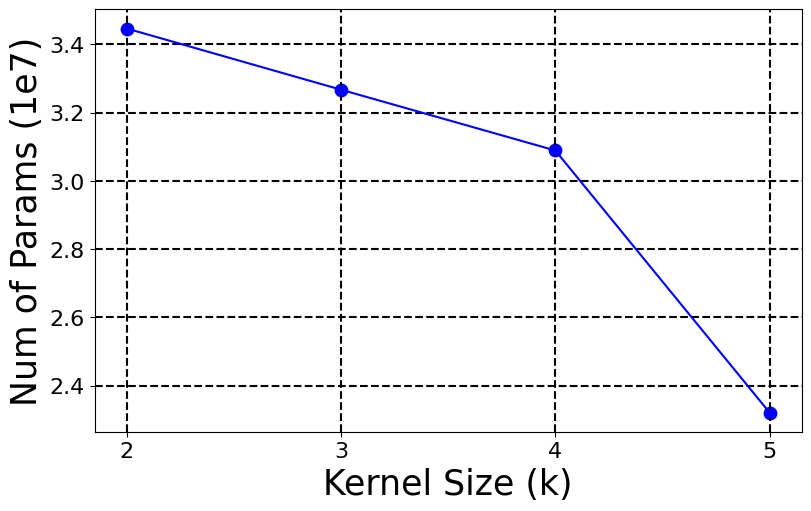}
        \caption{}
        \label{fig:vary_k_on_size}
    \end{subfigure}%
     ~
    \begin{subfigure}[t]{0.3\textwidth}
        \centering
        \includegraphics[width=1\textwidth]{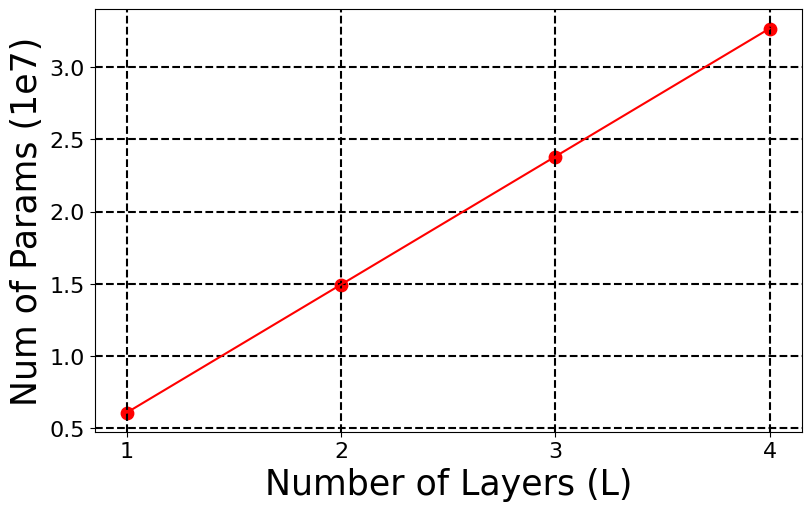}
        \caption{}
        \label{fig:vary_L_on_size}
    \end{subfigure}%
    ~
    \begin{subfigure}[t]{0.3\textwidth}
        \centering
        \includegraphics[width=1\textwidth]{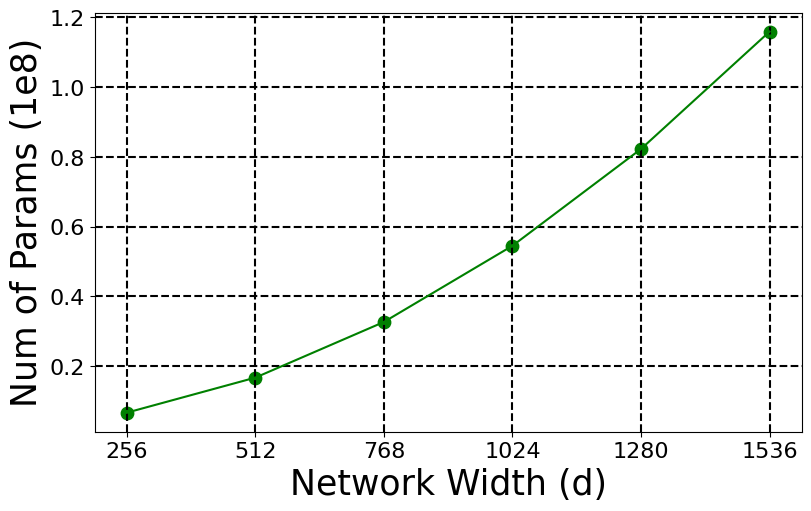}
        \caption{}
        \label{fig:vary_d_on_size}
    \end{subfigure}
    
    \caption{For the SEL task, effect on model size by (a)-(c): Varying the EDRNet's temporal kernel size $k$, number of (de/re)composition layers $L$ and network width $d$ respectively}
\end{figure*}

In the Ablation Studies, we observed the impact of the various EDRNet configurations on the performance on the SEL task. Here, we analyze how each of these dimensions affect the EDRNet's model size.

From Fig. \ref{fig:vary_k_on_size}, it is clear that on increasing the kernel size, the model size reduces. Since we do not use padding for the temporal convolutions, increasing $k$ limits the maximum number of decomposition and recomposition layers, hence causing the decreasing trend on the model size.  Fig. \ref{fig:vary_L_on_size} and \ref{fig:vary_d_on_size} reveals that increasing $L$ and $d$ increases the model size. While an increase in $L$ results in a linear increase in the model size, an increase in $d$ results in an exponential increase in the model size since almost every (de/re)composition operation in the network takes a $d$-dimensional input.

\subsection{Ablating Positional Encoding in the Event Decomposition Phase}

In the Event Decomposition Phase of the EDRNet, we add positional encoding (PE), computed with Equation \ref{eq:positional_encoding}, to the initial audio and visual features before the temporal convolutional layers (decomposition operations). As explained in the Event Recomposition Phase (ERP), the addition of the PE is critical to convey event-wise positional information to the localization branches of the ERP via the residual connections described in Equation \ref{eq:temporal_recomposition}. Without the PE, the decomposition operations temporally compresses the modality-wise video features, gradually eroding the events' position information within the video. Without the PE, the WSEL and SEL accuracy on the AVE dataset with the original test set degrades to 72.88\% (from 73.86\% with PE) ($\downarrow$0.98\%) and 77.09\% (from 77.93\% with PE) ($\downarrow$0.84\%) respectively.

\subsection{Event-Wise Performance Improvement from SMB Video Fusion}

\begin{figure*}[t!]
    \begin{subfigure}[t]{0.5\textwidth}
        \centering
        \includegraphics[width=1\textwidth]{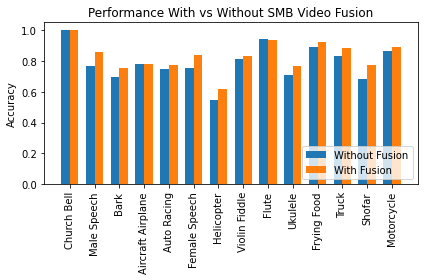}
    \end{subfigure}%
     ~
    \begin{subfigure}[t]{0.5\textwidth}
        \centering
        \includegraphics[width=1\textwidth]{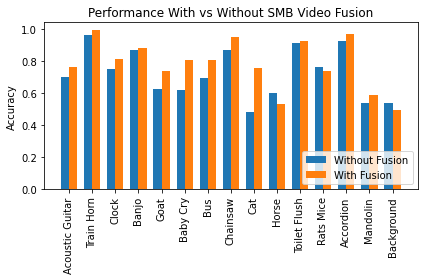}
    \end{subfigure}
\caption{Per event performance on the SEL task for EDRNet trained with vs without SMB fused videos.}
\label{fig:per_cat_with_without_smb}
\end{figure*}

In order to highlight the performance impact of the SMB video fusion in detail, we display the per-event category performance with and without the fused videos in Figure \ref{fig:per_cat_with_without_smb}. As observed, we consistently improve the SEL performance across most of the event categories. Background (BG) accuracy slightly decreases due to the BG-FG tradeoff. Additionally, we do not create SMB fused videos for the BG event since it is meaningless to semantically mix the numerous different situations which can encapsulate the broad concept of a BG event.

\section{Additional  Visualizations}

\subsection{Visualization of Dual-Phase Modality Fusion}

Although inspired from U-Net and ED-TCN, our EDRNet's unique structural adaptation contributes significantly to multi-modal temporal processing tasks where the timing and method of feature fusion becomes critical to the success of the network. As visualized in Figure \ref{fig:edrnet_dual_phase_fusion}, the EDRNet executes a novel \emph{dual-phase modality fusion} wherein the accumulation of individual modal perspectives at different temporal granularity during the decomposition phase progressively refines task-specific (e.g. localization) features during the recomposition phase. Results of the modal branch isolation experiments (Table \ref{tbl:modality_isolation_result}) prove the efficacy of the dual-phase fusion.

\subsection{Visualization of EDRNet for Modal Branch Isolation Experiments}

\begin{figure*}[t!]
    \centering
    \begin{subfigure}[t]{0.48\textwidth}
        \centering
        \includegraphics[width=1\textwidth]{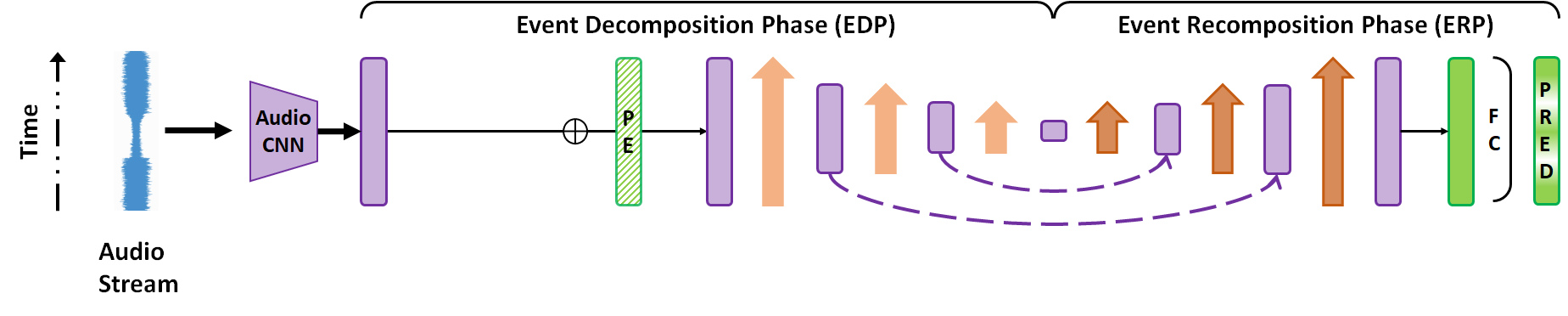}
        \caption{Audio-only activated decomposition branch of EDRNet. Corresponds to the 1st Row of Table \ref{tbl:modality_isolation_result}.}
    \end{subfigure}%
    ~
    \begin{subfigure}[t]{0.48\textwidth}
        \centering
        \includegraphics[width=1\textwidth]{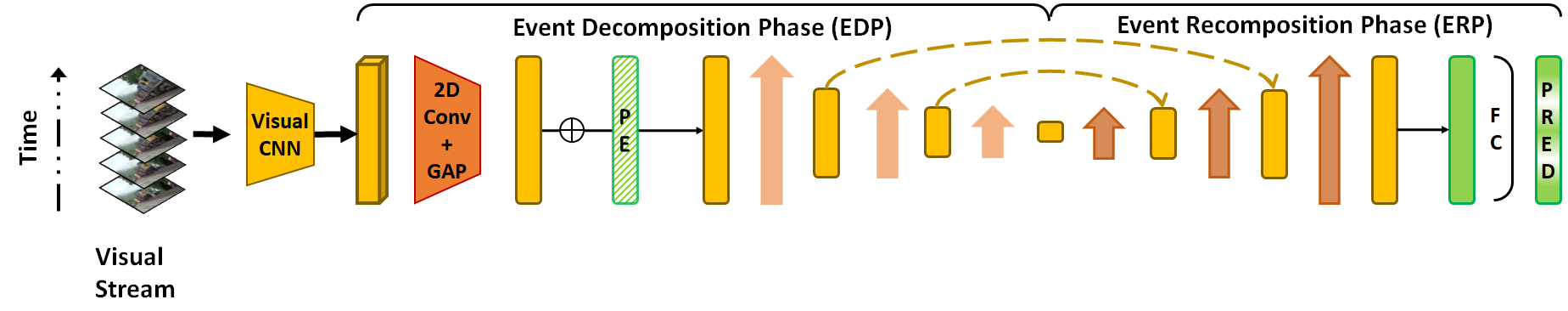}
        \caption{Visual-only activated decomposition branch of EDRNet. Corresponds to the 2nd Row of Table \ref{tbl:modality_isolation_result}.}
    \end{subfigure}
    
    \begin{subfigure}[t]{0.48\textwidth}
        \centering
        \includegraphics[width=1\textwidth]{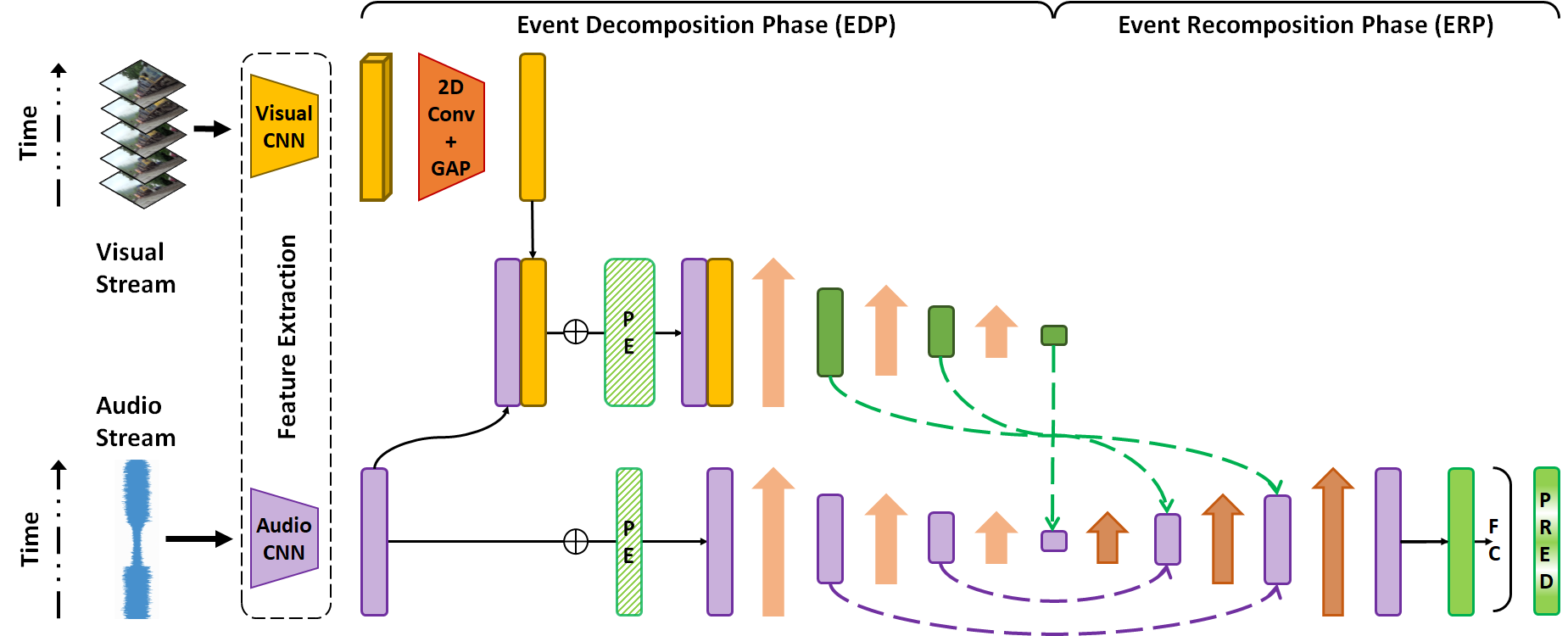}
        \caption{Audio and Dual Phase Fusion (AV) activated decomposition branches of EDRNet. Corresponds to the 3rd Row of Table \ref{tbl:modality_isolation_result}.}
    \end{subfigure}%
    ~
    \begin{subfigure}[t]{0.48\textwidth}
        \centering
        \includegraphics[width=1\textwidth]{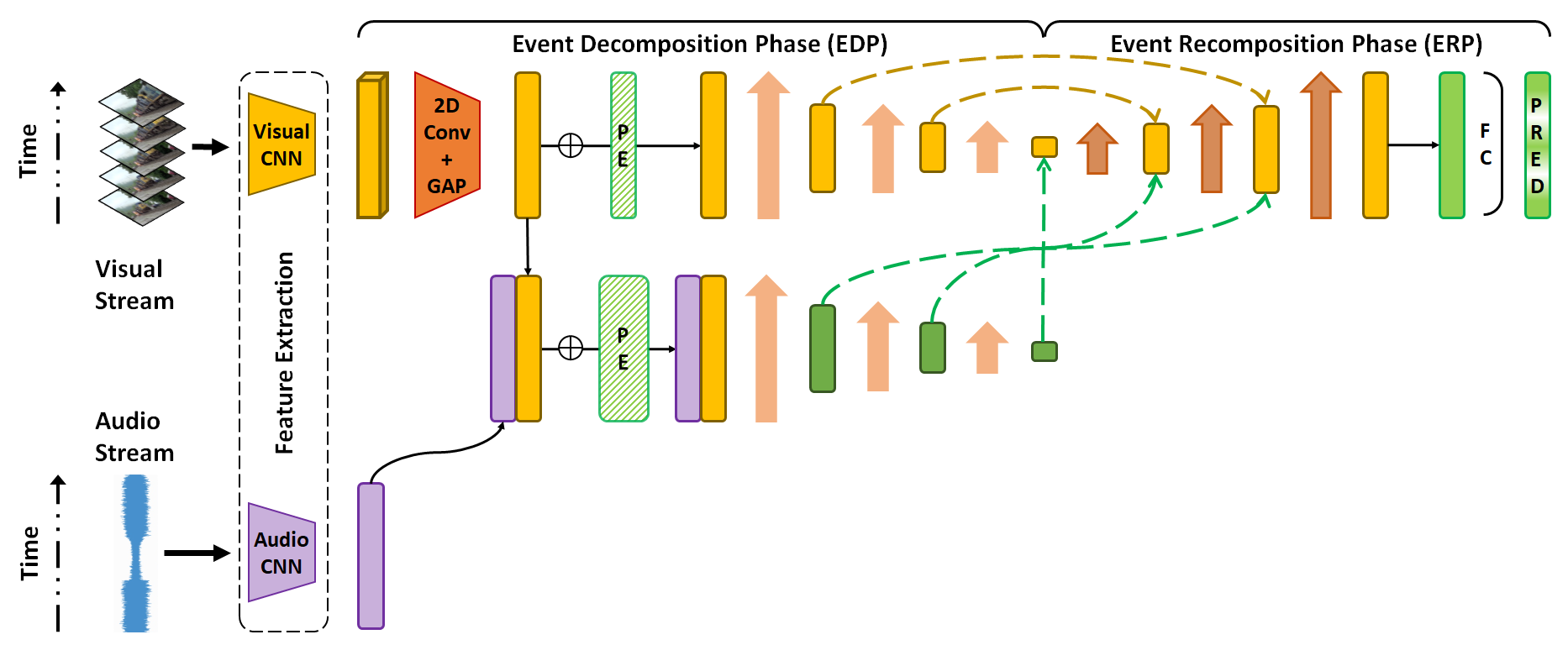}
        \caption{Visual and Dual Phase Fusion (AV) activated decomposition branches of EDRNet. Corresponds to the 4th Row of Table \ref{tbl:modality_isolation_result}.}
    \end{subfigure}

    \begin{subfigure}[t]{0.48\textwidth}
        \centering
        \includegraphics[width=1\textwidth]{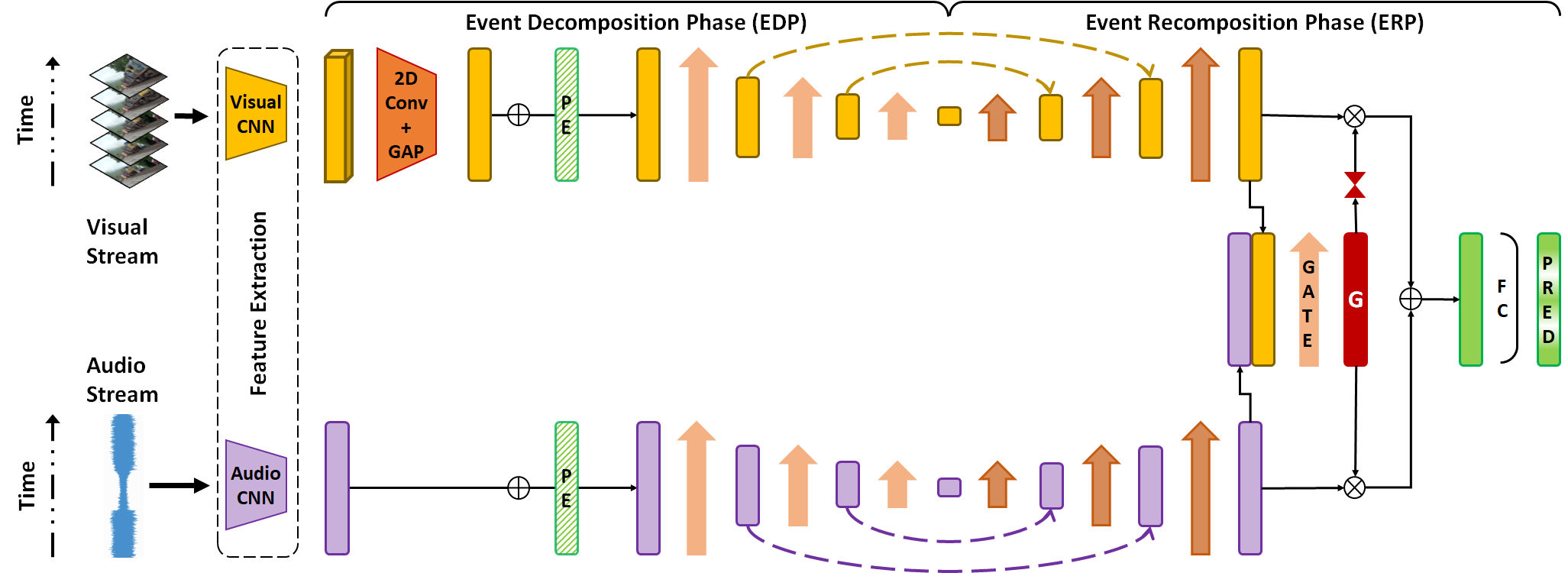}
        \caption{Audio and Visual activated decomposition branches of EDRNet, merged with late fusion. Corresponds to the 5th Row of Table \ref{tbl:modality_isolation_result}.}
    \end{subfigure}%
    ~
    \begin{subfigure}[t]{0.48\textwidth}
        \centering
        \includegraphics[width=1\textwidth]{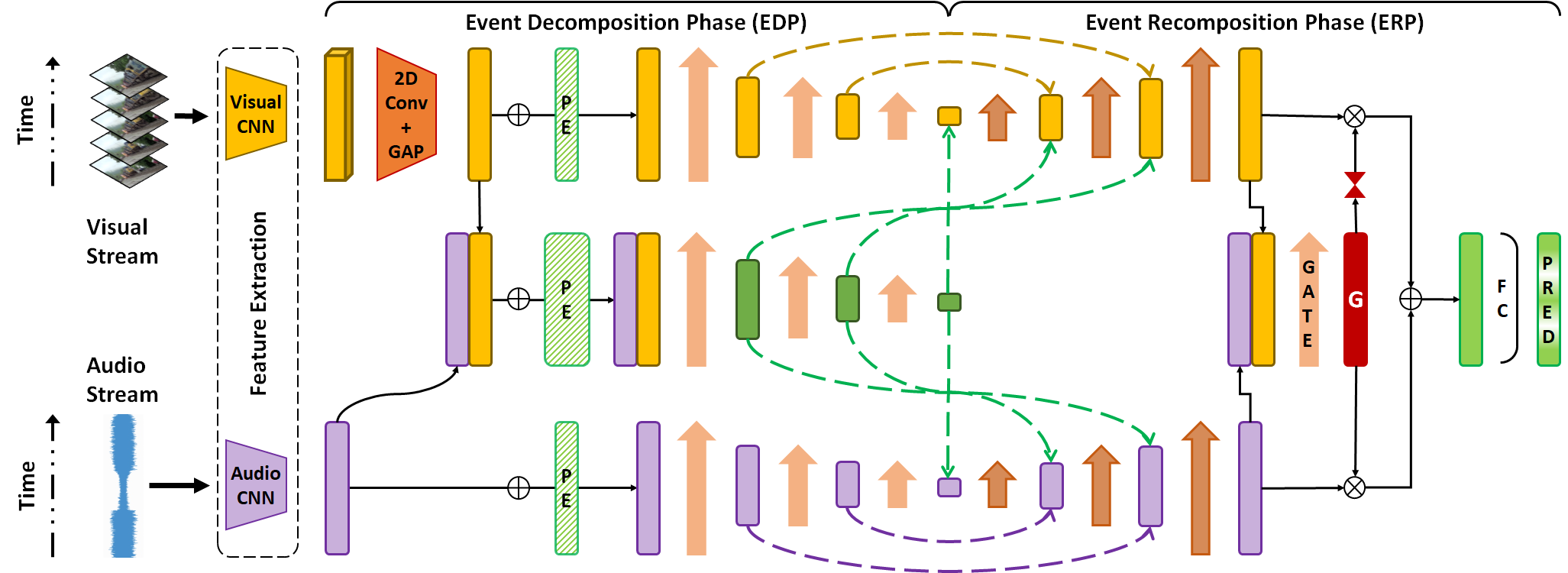}
        \caption{Audio, Visual and Dual Phase Fusion (AV) activated decomposition branches of EDRNet. Corresponds to the 6th Row of Table \ref{tbl:modality_isolation_result}.}
    \end{subfigure}
    \caption{Visualizations of EDRNet with various configurations corresponding to the modal branch isolation experiments.}
    \label{fig:supplem_edrnet_mb_isolation_exp}
\end{figure*}

In the last ablation study, we leveraged the structural assembly of the EDRNet to perform modal branch isolation experiments. We trained various combinations of the Event Decomposition Phase modal branches and studied the performance of the corresponding activated Event Recomposition
Phase branches. The training configurations and results are listed in Table \ref{tbl:modality_isolation_result}. To better articulate the experiments, we visualize EDRNet's architecture in Figure \ref{fig:supplem_edrnet_mb_isolation_exp} for each row of Table \ref{tbl:modality_isolation_result}.

\subsection{CAMs and SMB Fused Videos}
\begin{figure*}[t!]
    \centering
    \begin{subfigure}[t]{0.5\textwidth}
        \centering
        \includegraphics[width=1\textwidth]{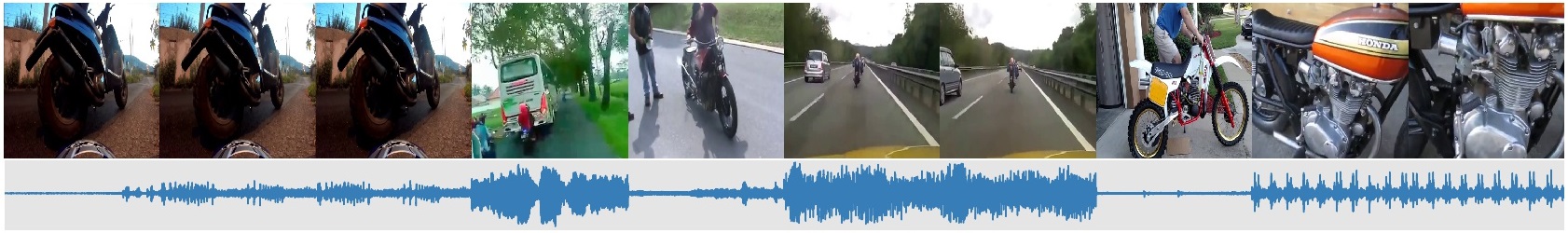}
        \caption{Event: Motorcycle}
    \end{subfigure}%
    ~
    \begin{subfigure}[t]{0.5\textwidth}
        \centering
        \includegraphics[width=1\textwidth]{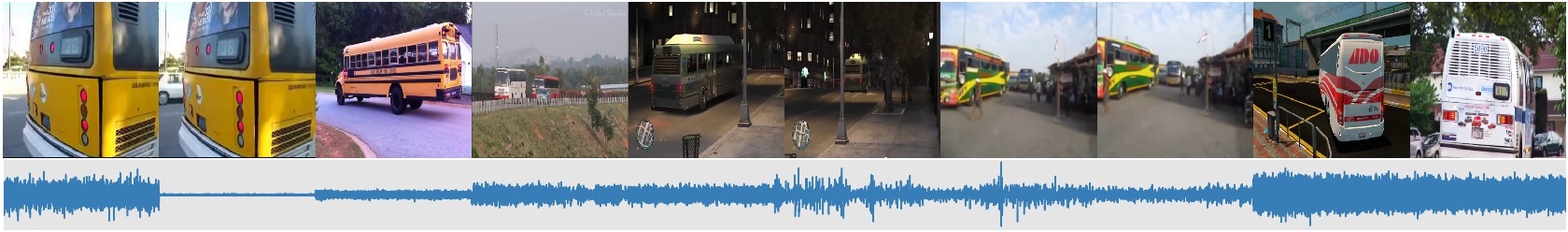}
        \caption{Event: Bus}
    \end{subfigure}
    \begin{subfigure}[t]{0.5\textwidth}
        \centering
        \includegraphics[width=1\textwidth]{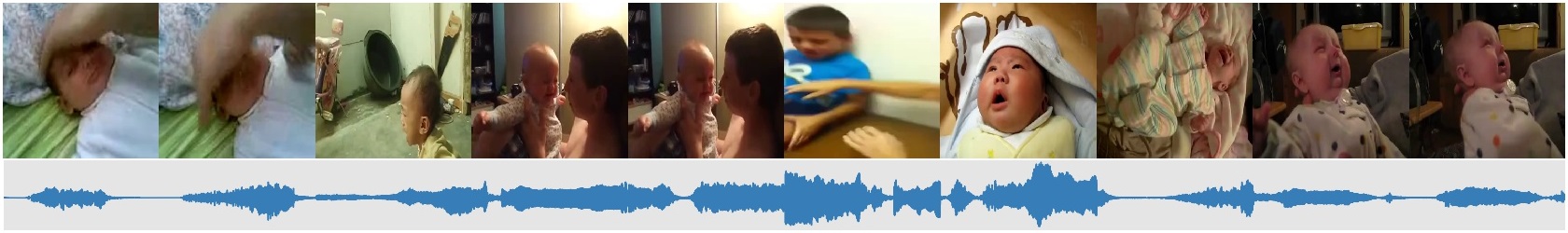}
        \caption{Event: Baby Crying}
    \end{subfigure}%
    ~
    \begin{subfigure}[t]{0.5\textwidth}
        \centering
        \includegraphics[width=1\textwidth]{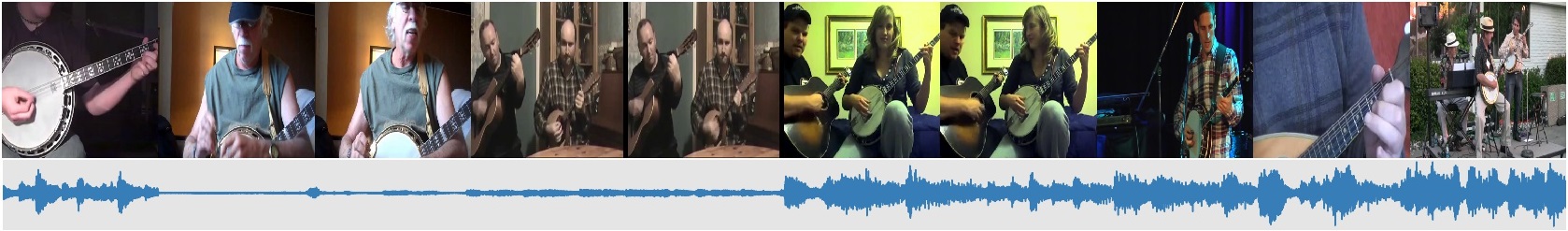}
        \caption{Event: Banjo}
    \end{subfigure}    
    \begin{subfigure}[t]{0.5\textwidth}
        \centering
        \includegraphics[width=1\textwidth]{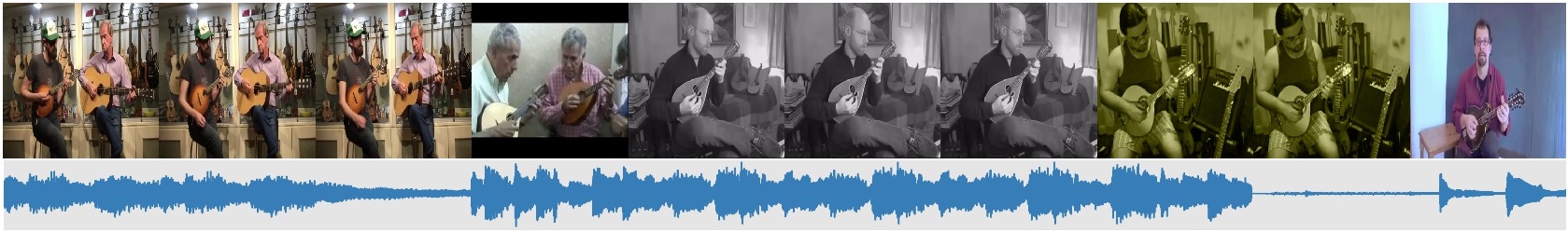}
        \caption{Event: Guitar}
    \end{subfigure}%
    ~
    \begin{subfigure}[t]{0.5\textwidth}
        \centering
        \includegraphics[width=1\textwidth]{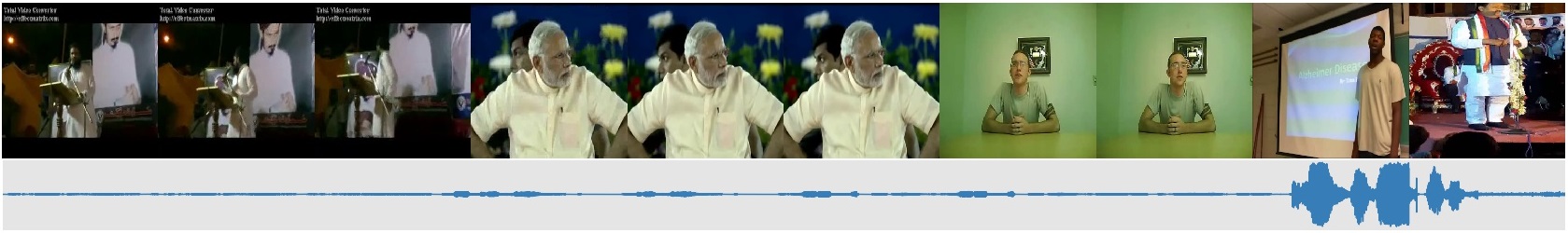}
        \caption{Event: Male Speech}
    \end{subfigure}
    \begin{subfigure}[t]{0.5\textwidth}
        \centering
        \includegraphics[width=1\textwidth]{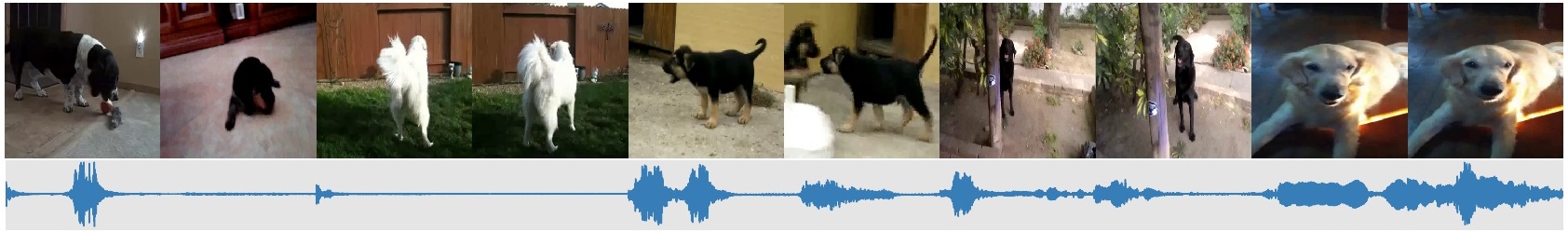}
        \caption{Event: Dog Barking}
    \end{subfigure}%
    ~
    \begin{subfigure}[t]{0.5\textwidth}
        \centering
        \includegraphics[width=1\textwidth]{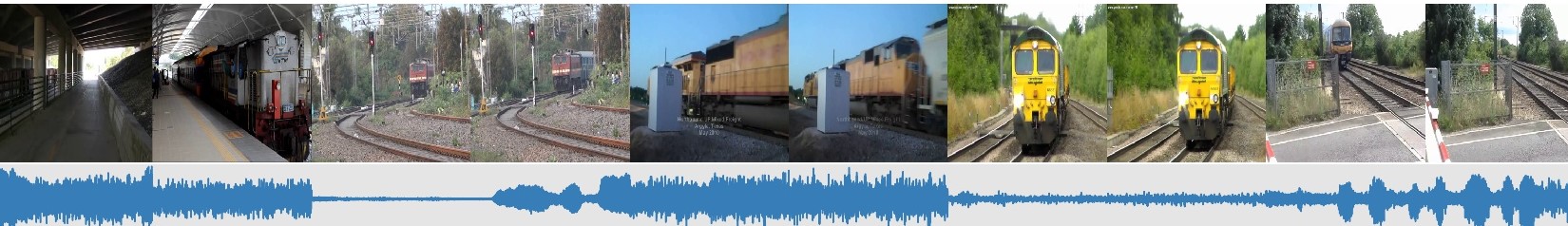}
        \caption{Event: Train}
    \end{subfigure}
    \caption{SMB fused videos for various events}
    \label{fig:supplem_smb_vis}
\end{figure*}

\begin{figure*}[t!]
    \centering
    \begin{subfigure}[t]{0.5\textwidth}
        \centering
        \includegraphics[width=1\textwidth]{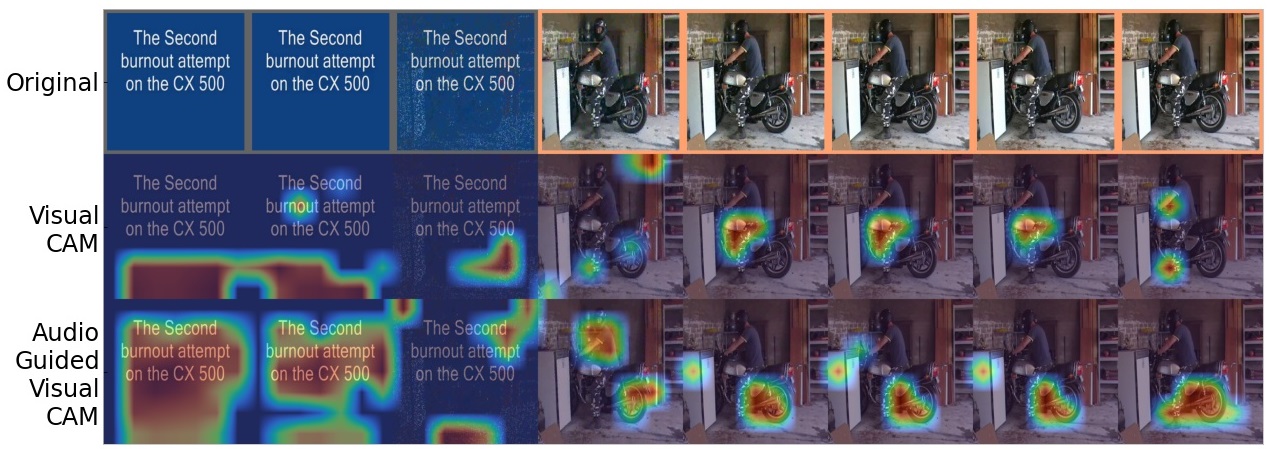}
        \caption{Event: Motorcycle}
    \end{subfigure}%
    ~
    \begin{subfigure}[t]{0.5\textwidth}
        \centering
        \includegraphics[width=1\textwidth]{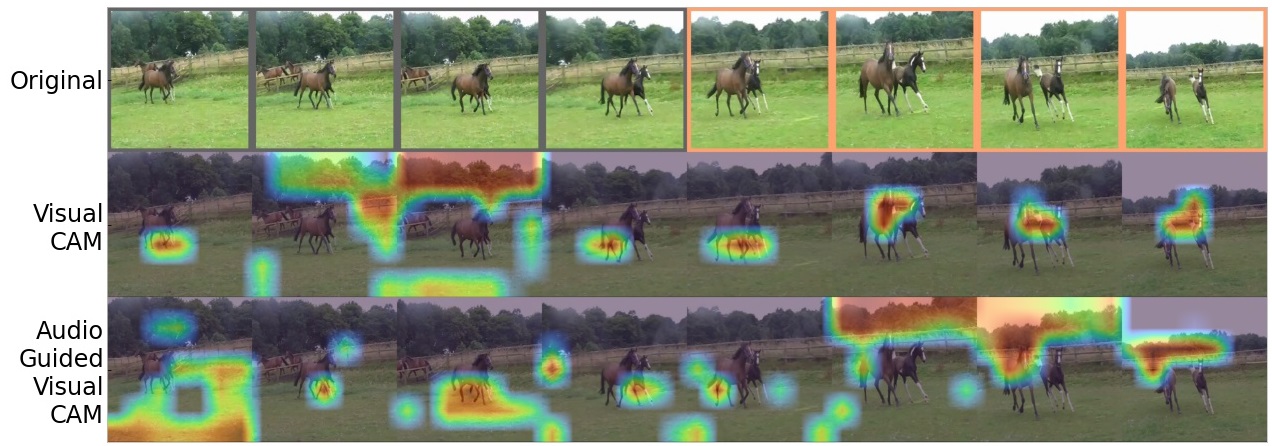}
        \caption{Event: Horse}
    \end{subfigure}
    \begin{subfigure}[t]{0.5\textwidth}
        \centering
        \includegraphics[width=1\textwidth]{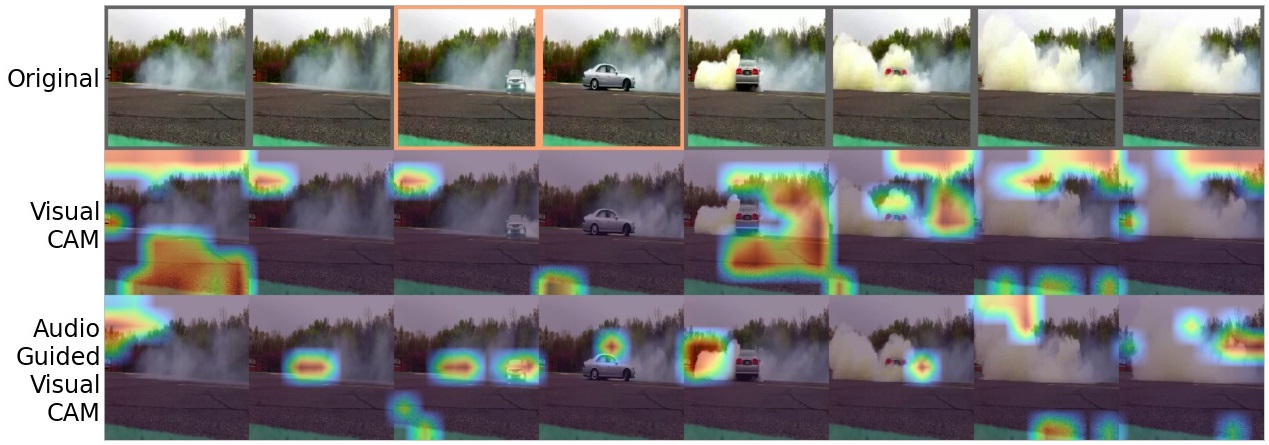}
        \caption{Event: Racecar}
    \end{subfigure}%
    ~
    \begin{subfigure}[t]{0.5\textwidth}
        \centering
        \includegraphics[width=1\textwidth]{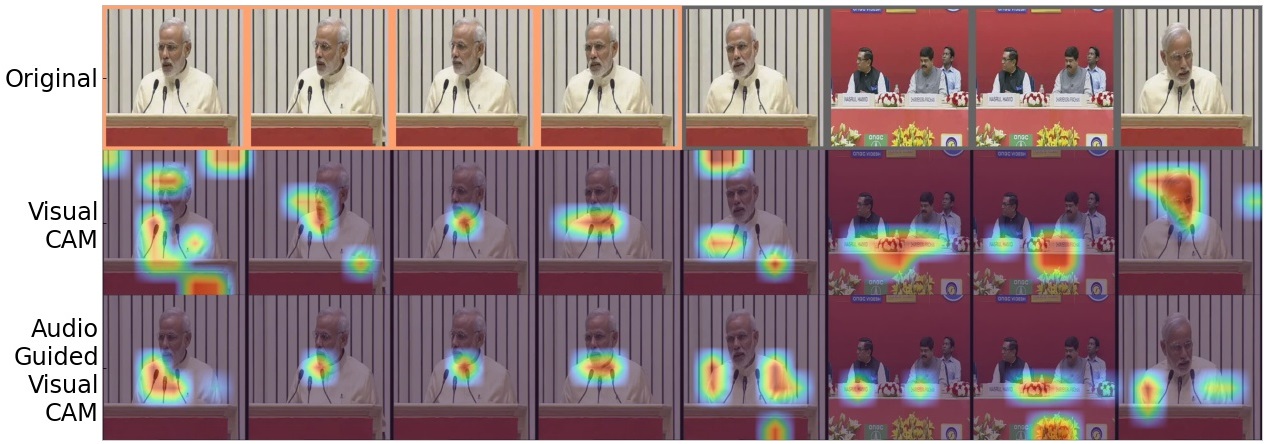}
        \caption{Event: Male Speech}
    \end{subfigure}    
    \begin{subfigure}[t]{0.5\textwidth}
        \centering
        \includegraphics[width=1\textwidth]{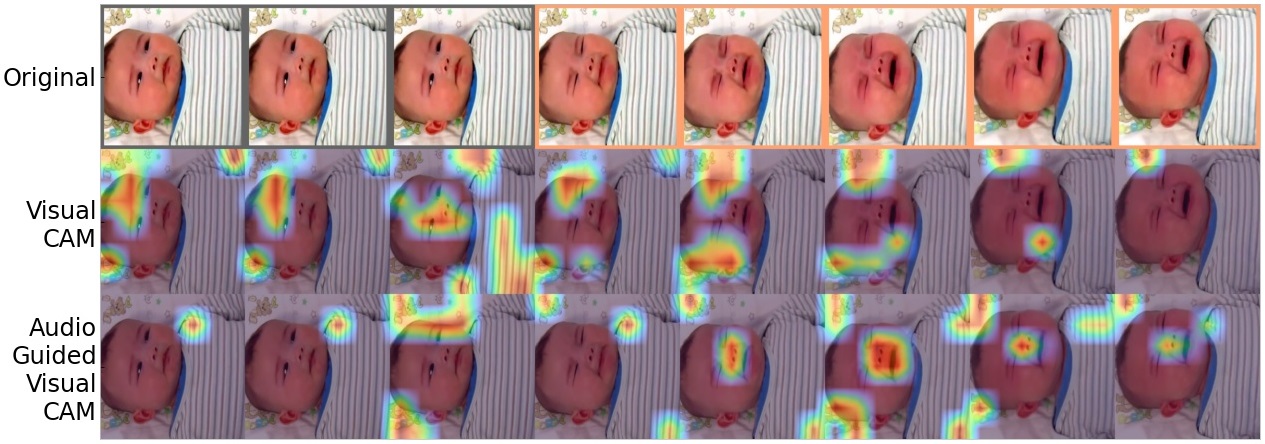}
        \caption{Event: Baby Crying}
    \end{subfigure}%
    ~
    \begin{subfigure}[t]{0.5\textwidth}
        \centering
        \includegraphics[width=1\textwidth]{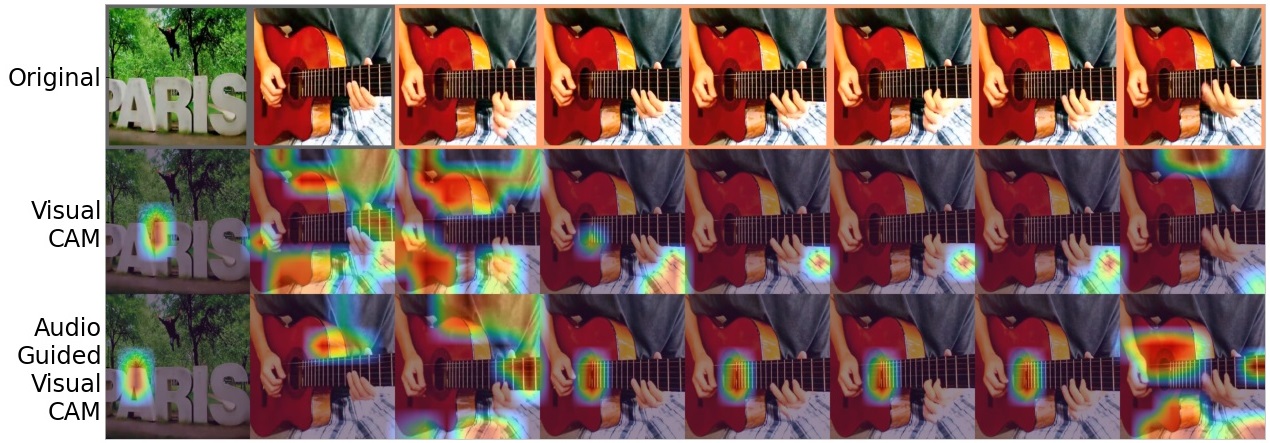}
        \caption{Event: Playing Guitar}
    \end{subfigure}
    \begin{subfigure}[t]{0.5\textwidth}
        \centering
        \includegraphics[width=1\textwidth]{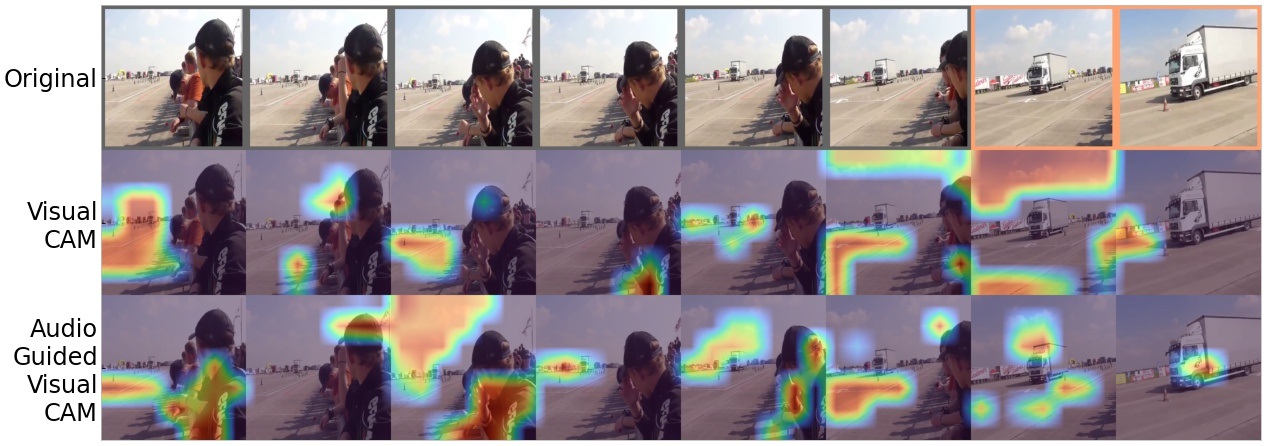}
        \caption{Event: Truck}
    \end{subfigure}%
    ~
    \begin{subfigure}[t]{0.5\textwidth}
        \centering
        \includegraphics[width=1\textwidth]{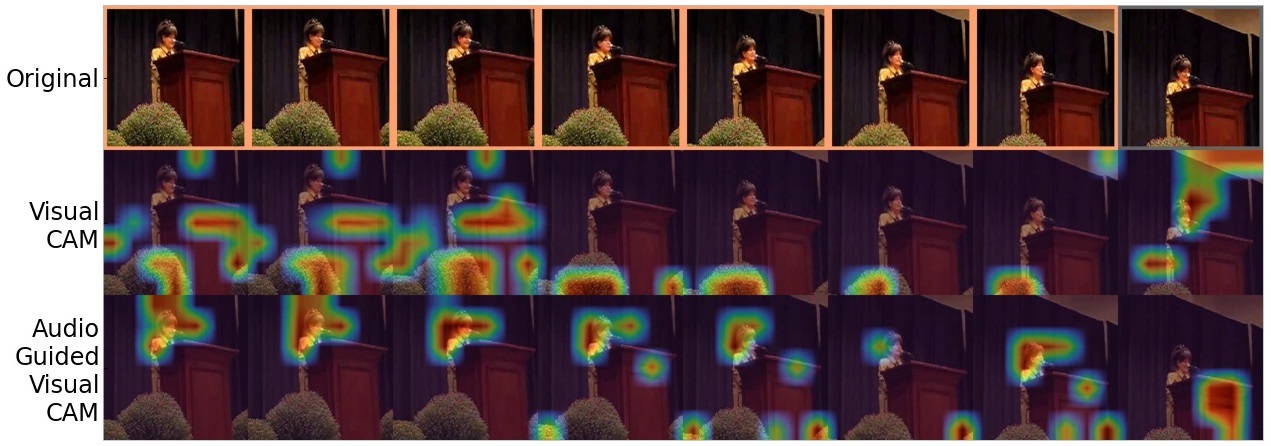}
        \caption{Event: Female Speech}
    \end{subfigure}
    \caption{CAM visualizations for various events. Orange bordered segments indicate the presence of FG event.}
    \label{fig:supplem_cam_vis}
\end{figure*}

In Figure \ref{fig:supplem_smb_vis}, we provide audio-visual sequence visualizations of a few generated videos using the SMB video fusion technique. Additionally, in Figure \ref{fig:supplem_cam_vis}, we provide additional CAM visualizations for videos of various events to demonstrate the on-demand cross-modal guidance achieved by the EDRNet.

\clearpage

\end{document}